\documentclass[10pt,journal,compsoc]{IEEEtran}

\usepackage{gensymb}
\usepackage{amsmath}
\usepackage[table]{xcolor}
\usepackage{hhline}
\usepackage{svg}
\usepackage{float}
\usepackage{graphics}
\usepackage{booktabs}

\ifCLASSOPTIONcompsoc
  \usepackage[nocompress]{cite}
\else
  \usepackage{cite}
\fi
\ifCLASSINFOpdf
\else
\fi
\begin{document}
%
\title{Full Body Video-Based Self-Avatars for Mixed Reality: from E2E System to User Study}
%
%
%
%

\author{Diego Gonzalez Morin,
        Ester Gonzalez-Sosa,
        Pablo Perez,
        and Alvaro Villegas
\IEEEcompsocitemizethanks{\IEEEcompsocthanksitem Diego Gonzalez-Morin and Ester Gonzalez-Sosa contribute equally\protect\\
diego.gonzalez\_morin@nokia.com \\
ester.gonzalez@nokia.com \\
pablo.perez@nokia.com \\
alvaro.villegas@nokia.com \\
}
}

%
%

\markboth{Journal of Transactions on Visualization and Computer Graphics}%
{Shell \MakeLowercase{\textit{et al.}}: Bare Demo of IEEEtran.cls for Computer Society Journals}
%


\IEEEtitleabstractindextext{%
\begin{abstract}
In this work we explore the creation of self-avatars through video pass-through in Mixed Reality (MR) applications. We present our end-to-end system, including: custom MR video pass-through implementation on a commercial head mounted display (HMD), our deep learning-based real-time egocentric body segmentation algorithm, and our optimized offloading architecture, to communicate the segmentation server with the HMD. To validate this technology, we designed an immersive VR experience where the user has to walk through a narrow tiles path over an active volcano crater. The study was performed under three body representation conditions: virtual hands, video pass-through with color-based full-body segmentation and video pass-through with deep learning full-body segmentation. This immersive experience was carried out by 30 women and 28 men. To the best of our knowledge, this is the first user study focused on evaluating video-based self-avatars to represent the user in a MR scene. Results showed no significant differences between the different body representations in terms of presence, with moderate improvements in some Embodiment components between the virtual hands and full-body representations. Visual Quality results showed better results from the deep-learning algorithms in terms of the whole body perception and overall segmentation quality. We provide some discussion regarding the use of video-based self-avatars, and some reflections on the evaluation methodology. The proposed E2E solution is in the boundary of the state of the art, so there is still room for improvement before it reaches maturity. However, this solution serves as a crucial starting point for novel MR distributed solutions. 
\end{abstract}
}

\maketitle
\IEEEdisplaynontitleabstractindextext

%
\IEEEpeerreviewmaketitle
\IEEEraisesectionheading{\section{Introduction}\label{sec:introduction}}

%
%
%
%


\IEEEPARstart{T}{he} use of self-avatars, the virtual representation of the user's body from its own perspective, is starting to become ubiquitous in Virtual Reality (VR) and Mixed Reality (MR). The possibility of seeing your own body while immersed brings many benefits. First, it increases the users' Sense of Presence (SoP), as it shifts the user from being an  observer to really experiencing the Virtual Environment (VE) \cite{lok2003effects,slater1993influence}. Besides, it enhances spatial perception and distance estimation \cite{mcmanus2011influence,ebrahimi2018investigating} while bringing a positive impact on cognitive load \cite{steed2016impactselfavatar}, trust and collaboration \cite{pan2017impact}. Apart from increasing the SoP, self-avatars also increase the Sense of Embodiment (SoE) \cite{argelaguet2016role,fribourg2020avatar}. As defined by Kilteni \textit{et al.} \cite{kilteni2012sense}, SoE refers to the feeling of being inside, controlling and having a virtual body. SoE includes the sense of \textit{ownership} as one's self attribution of a body, sense of \textit{location} defined as one's spatial experience of being inside a body, and sense of \textit{agency} as the sense of having global motor control over a body.

\begin{figure}
  \centering
  \includegraphics[width=1.0\linewidth]{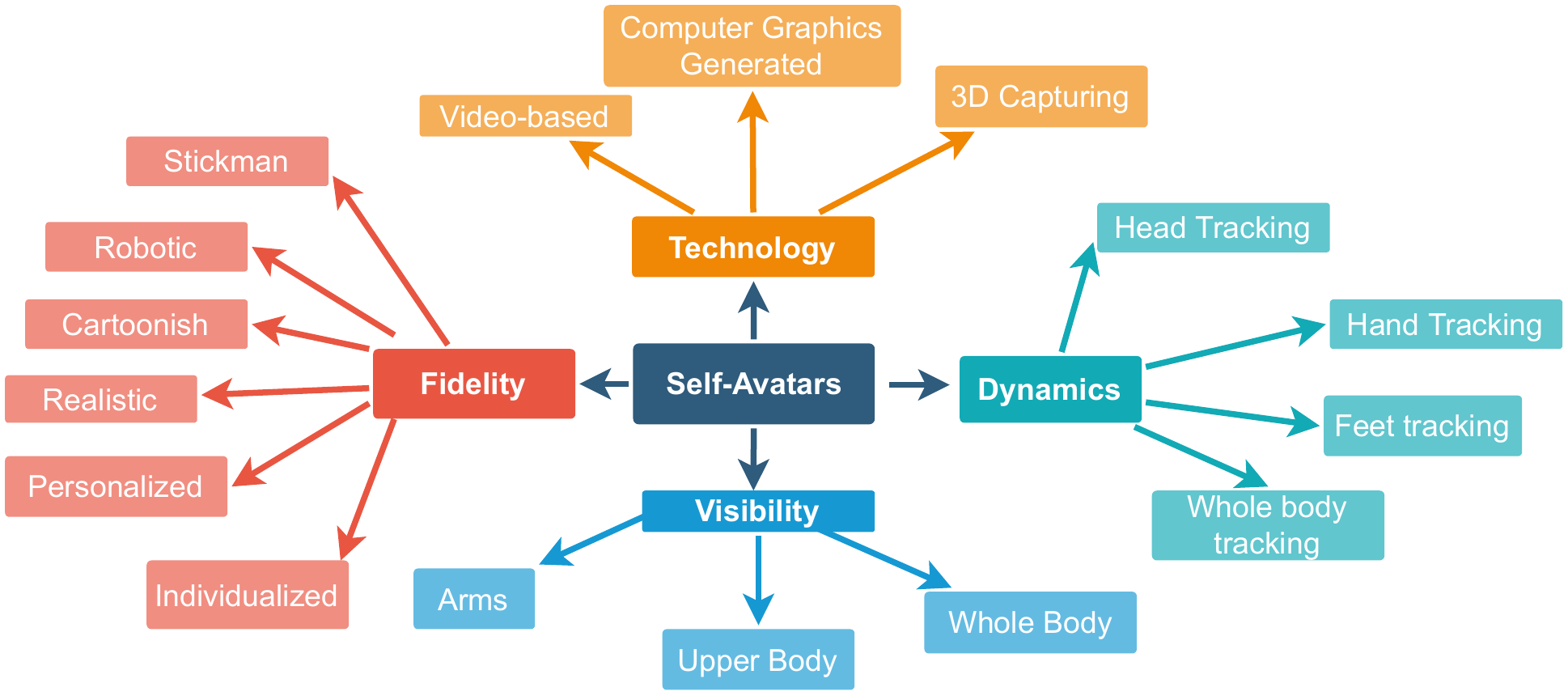}
  \caption{Self-avatars can be described based on their visibility, fidelity and dynamics. There are different related technologies}
  \label{fig:selfavatar}
\end{figure}

In a nutshell, an avatar is a user representation that can be described based on its visibility, fidelity and dynamics (see Fig.\ref{fig:selfavatar}). In what concerns to avatars visibility, the minimal configuration of an avatar just relies on modelling the users' hands \cite{dodds2011talk}. On the other hand, the current trend is to model the partial body (e.g., upper body limbs \cite{gruosso2021}) or even the entire human body \cite{ogawa2020you}.

Once decided the level of visibility, it is necessary to decide which avatar fidelity or realism level (shape and texture for skin and clothes), should be used for modelling the human body. Existing works for hand representations range from abstract forms \cite{argelaguet2016role}, minimal iconic representation \cite{argelaguet2016role}, robot hands, to more realistic ones with gender-matched and skin-matched limbs \cite{dewez2019ismar}. Fidelity approaches for full body avatars includes minimal representations, such as the stickman appearance \cite{fribourg2020avatar}, cartoonish ones (e.g., the ones used in Mozilla Hubs), realistic avatars usually found in  human model libraries such as Rocketbox library \cite{gonzalez2020rocketbox}, or personalized ones found in VR platforms (e.g. Virbella). The latest tendency relies on models that preserve the user's body shape such as using a skinned multi-person linear body (SMPL) model \cite{loper2015smpl}, or 3D scanning systems such as point-cloud representation for fully personalized avatars \cite{thaler2018visual}. There is still room for improvement with modelling identity-preserving avatars, specially concerning real-time performance issues. It remains an interesting challenge, since it has been proven the benefits of avatar fidelity and realism level in object size perception \cite{ogawa2019virtual,ogawa2020you}, body ownership \cite{gorisse2019robot}, presence \cite{waltemate2018impact}, co-presence \cite{yu2021avatars}, among others. 

One remaining challenge is how to capture all movements from the user's body so that the avatar can accurately mimic them. It is an important feature as it facilitates interaction tasks in MR \cite{pan2019foot}. Modelling head and hands movements have been widely studied. Head tracking is provided by almost all standard HMDs. Besides, there are recent commercial solutions for hand tracking, such as Leap Motion sensor \cite{serra2020natural}, or camera-based hand tracking as in the Meta Quest 2 or HTC VIVE. On the other hand, less work has been done on full body tracking, including lower limbs and torso. Some works proposed using 2 VIVE trackers attached to the feet for allowing foot tracking and visibility \cite{pan2019foot,bonfert2022kicking,bozgeyikli2022tangiball}. Although traditional approaches for full body capture are based on inertial measurements units (IMUs) sensors placed on a suit that the user wears \cite{xu2019towards,waltemate2018impact,jayaraj2017improving}, there are emerging solutions based on computer vision that can track the user's body movements using one single depth sensor such as Azure Kinect v2 \cite{gonzalez2020movebox}. Apart from the already mentioned challenges of full body tracking and avatar appearance resembling the user's own identity, there is still the subjective factor of whether users accept the avatar as a virtual representation of themselves or not.


An alternative approach to computer-generated (CG) avatars is the use of video-based self-avatars. This can be done through the combination of VR video pass-through and computer vision algorithms: user's real body is incorporated into the virtual scene by segmenting the egocentric vision of a stereo camera placed in front of or integrated with the HMD. The VR community has explored this idea over the last decade. For instance color-based approaches \cite{fiore2012towards,bruder2009enhancing,immersirve_gastronomic2019,gunther2015aughanded} can be deployed in real time but they tend to work well just for controlled conditions. Further, they failed at dealing with different skin colors or long-sleeve clothes \cite{fiore2012towards}. Alternatively, depth cameras solutions have been proposed \cite{rauter2019augmenting,lee2016enhancing,alaee2018user}. Despite being effective for some situations, they still lack precision to provide a generic, realistic, and real time immersive experience. In our previous work \cite{gonzalez2020enhanced,gonzalez2022_realtimeseg} we proposed a real-time egocentric body segmentation algorithm. This algorithm, inspired in Thundernet’s architecture \cite{xiang2019thundernet}, achieves a frame rate of 66 fps for an input resolution of $640x480$, thanks to its shallow design. To train it, we created a $10,000$ images dataset with high variability in terms of users, scenarios, illumination, gender, among others. Therefore, this paper extends our previous work  \cite{gonzalez2020enhanced,gonzalez2022_realtimeseg} \cite{gonzalez2022bringing,morin2022democratic} and further investigates how to integrate the real time egocentric user body segmentation algorithm. Our main contributions can be summarized as: 
\begin{itemize}
    \item a detailed description of our E2E system which manages to integrate our real time egocentric segmentation algorithm in a realistic MR experience. It is composed of three main modules: video pass-through capable VR device, real-time egocentric body segmentation algorithm, and optimized offloading architecture.
    \item a subjective evaluation of the video-based self-avatar technology integrated in a gamified immersive experience, conducted by $58$ users. To the best of our knowledge, it is the first work that includes a user study using full body video-based self-avatars.
\end{itemize}

The rest of this article is structured as follows: Section~\ref{sec:related} covers relevant state of the art regarding the use of segmentation algorithms in Augmented Virtuality (AV) and Augmented Reality (AR) domains. Section \ref{sec:integration} describes the implementation details of our egocentric body segmentation E2E system. Section \ref{sec:immersive_experience} gives detail about the gamified immersive experience designed to validate the video-based self-avatar technology. Section \ref{sec:results} presents the subjective results obtained with a set of $58$ different subjects performing the immersive experience. We describe the conducted between-subjects and within-subjects experiments. Finally Section \ref{sec:discu} further elaborates on the benefits and drawbacks of using video-based avatars in place of CGI avatars. Last Section \ref{sec:conclu} concludes the paper. 


\begin{figure}
  \centering
  \includegraphics[width=1.0\linewidth]{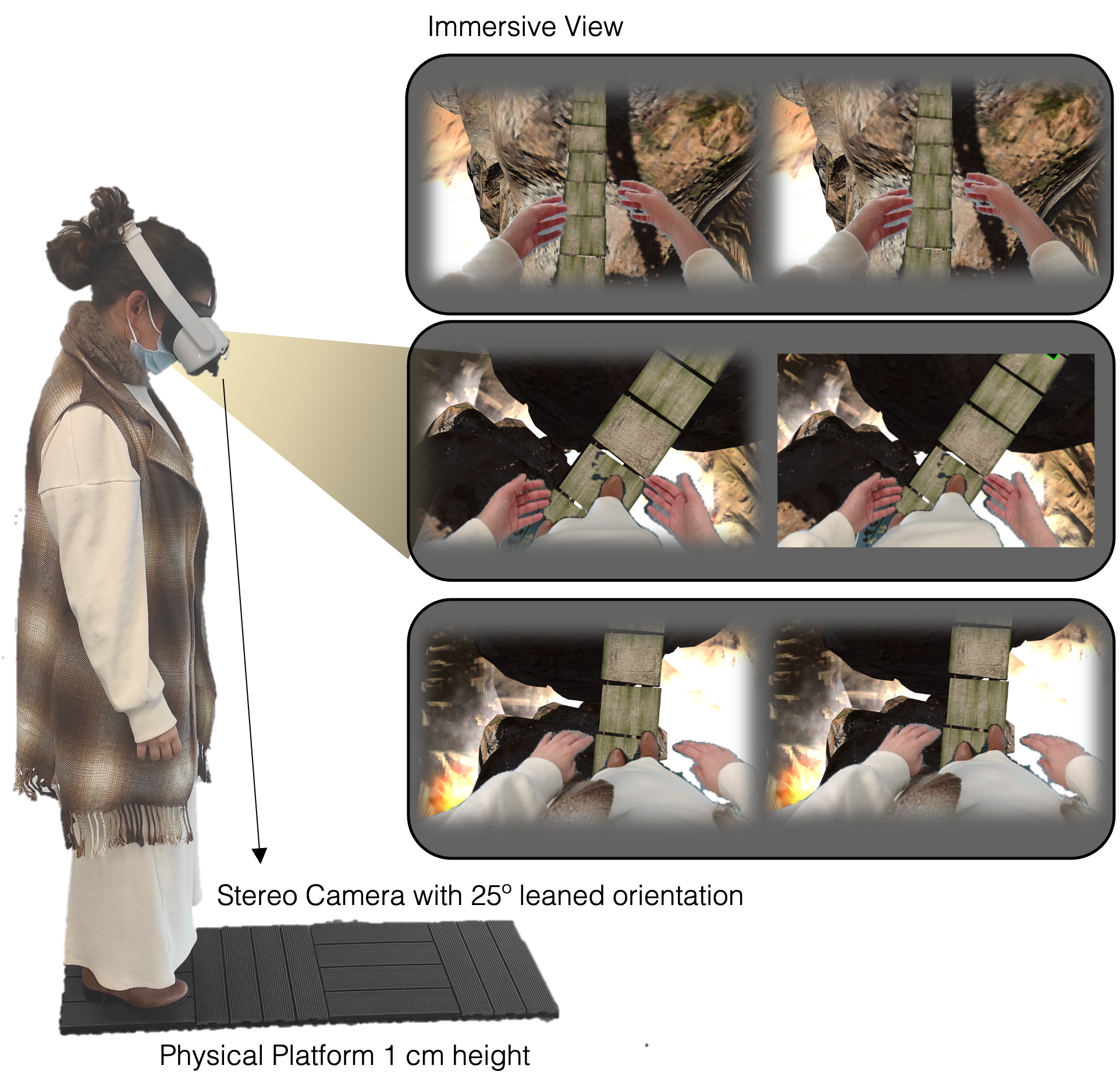}
  \caption{Example of video-based self-avatar in an immersive experience.}
  \label{fig:Cameras}
\end{figure}

\section{Related Works}
\label{sec:related}

\subsection{Augmented Virtuality}

\begin{figure*}
  \centering
   \includegraphics[width=0.9\linewidth]{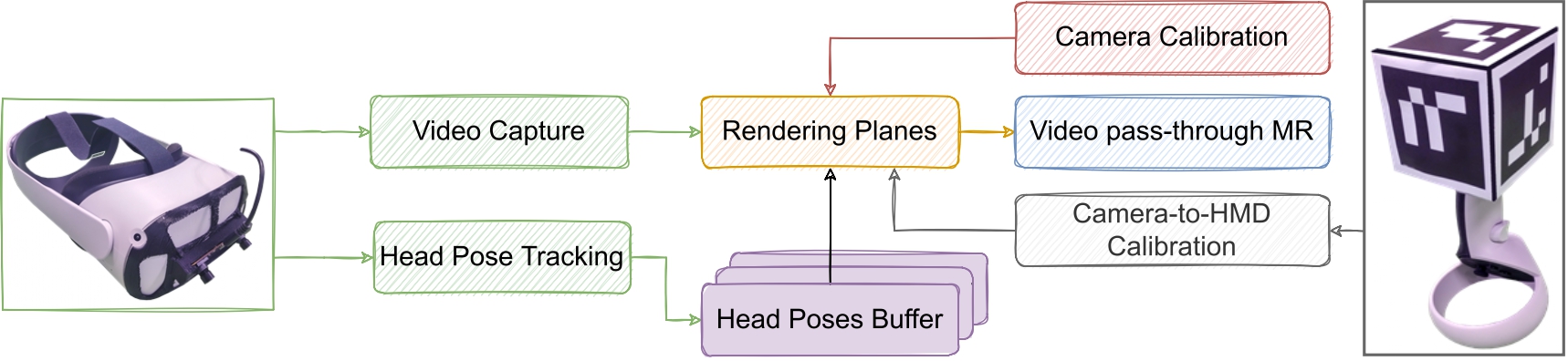}
  \caption{Simplified representation of our custom video pass-through implementations. }
  \label{fig:passthru}
\end{figure*}

Augmented Virtuality is a MR subcategory of the reality continuum that aims to merge portions of the reality surrounding the user with a virtual environment. In particular, introducing user's own body has attracted a lot of interest to the scientific community. For instance, Bruder \textit{et al.}~proposed skin segmentation algorithm to incorporate users' hands handling different skin colors \cite{bruder2009enhancing}. Conversely, a floor subtraction approach was developed to incorporate users' legs and feet in the VE. Assuming that the floor appearance was uniform, the body was retained by simply filtering out all pixels not belonging to the floor \cite{bruder2009enhancing}. 

Chen \textit{et al.}, in the context of 360$^{\circ}$ video cinematic experiences, explored depth keying techniques to incorporate all objects below a predefined distance threshold \cite{chen2017effect}.~This distance threshold could be changed dynamically to control how much of the real world was shown in the VE. The user could also control the transitions between VE and real world through head shaking and hand gestures. Some of the limitations that the authors pointed out were related to the reduced field of view of the depth sensor.

Pigny et Dominjon \cite{pigny2019using} were the first to propose a deep learning algorithm to segment egocentric bodies. Their architecture was based on U-NET and it was trained using a hybrid dataset composed of images from the COCO dataset belonging to persons and a $1500$-image custom dataset created following the automatic labelling procedure reported in \cite{gonzalez2020enhanced}. The segmentation network has two decoders: one for segmenting egocentric users, the other for exocentric users. However their segmentation execution speed could not currently totally keep up with the cameras update rate while maintaining a decent quality. They reported $16$ ms of inference time for $256\times256$ images, and observed problems of false positives that downgrade the AV experience.

Later, in \cite{gruosso2021}, the authors developed a system able to segment the user's hand and arms using a dataset composed of $43K$ images from TeGO, EDSH, and a custom EgoCam dataset, using semantic segmentation deep learning network based on DeepLabv3+. For the experiment, they used an HTC Vive HMD, a monocular RGB camera, and a Leap Motion controller to provide free-hands interaction camera. Inference time takes around $20$ms for $360x360$ RGB video in a workstation provided with Nvidia Titan Xp GPU (12GB memory). They manage to run VR rendering and CNN segmentation in the same workstation.

In the last years, there is an increase interest in bringing real objects beyond human body parts to the VE. Preliminary works can be found in \cite{bai2021bringing}, where the mobile phone is segmented using VR controllers, in \cite{cortes2022qoe} where training tools are segmented with green chroma, or in \cite{tian2019enhancing}, where objects are segmented from bounding boxes previously estimated using standard object recognition networks.

\subsection{Augmented Reality}

In the context of Augmented Reality, real-time semantic segmentation is used to allow the dynamic and realistic occlusion of virtual objects rendered on top of the real scene. This feature is extremely relevant in AR as it allows the user to correctly perceive the 3D position of the virtual objects within the real-world. For example, if the user places his/her hand in front of a virtual object the hand should occlude it, otherwise the object would be perceived as being in front of the user's hand. Besides, the semantic information can be used to blend virtual objects interacting with real hands or other real objects in a more coherent and realistic way. More recent approaches combine the use of semantic information and inaccurate depth precision maps to provide a more realistic blending between virtual and foreground objects \cite{roxas2018occlusion}.
 

\section{SYSTEM DESIGN AND IMPLEMENTATION}

The generation of real-time, accurate and effective video-based self-avatars for MR requires the following solutions to be integrated as a single MR end-to-end system: 
\begin{itemize}
    \item \textbf{Video pass-through capable VR device}: captures the real scenario using a frontal stereo camera, accurately aligns it to the user's view and head movement, and renders it, in real-time, within the immersive scene. 
    \item \textbf{Real-time egocentric body segmentation algorithm}: identifies the pixels corresponding to the user's body from the frames captured by the frontal camera. 
    \item \textbf{Optimized offloading architecture}: egocentric body segmentation algorithms require high-end hardware not available in current state of the art HMDs. Consequently, we need a communication architecture which allows the fast data exchange between the immersive client and the server running the segmentation algorithm. 
\end{itemize}

\label{sec:integration}

\subsection{Custom Video pass-through Implementation}

There are several VR devices with video pass-through capabilities which are already commercially available, such as the Varjo XR-3\footnote{https://varjo.com/products/xr-3/} or the HTC VIVE Pro\footnote{https://www.vive.com/eu/product/vive-pro/}. These devices are still tethered, constraining the range of possible use cases of the proposed system. Consequently, we decided to build our own video pass-through solution \cite{morin2022democratic} for the Meta Quest 2\footnote{https://store.facebook.com/en/quest/products/quest-2/}, as it is a well-known commercially successful stand alone VR device. For this purpose, we had to build our own hardware along with the necessary software to effectively integrate the captured video into the VR scene, ensuring a perfect alignment between the captured feed and head movement. 

\subsubsection{Stereo Camera and 3D-printed Attachment}

Our video pass-through system must capture high definition stereoscopic video at a rate of at least 60 Hz to comply with many MR video pass-through use cases\footnote{https://developer.oculus.com/resources/oculus-device-specs/}. We chose the ELP 960P HD OV9750 stereo camera\footnote{https://bit.ly/3qGHY8h} which provides a maximum resolution of 2560x960 at a 60Hz update rate and a field of view of $90\degree$. Most of commercially available video pass-through MR devices incorporate stereo cameras which are aligned with the user's eyes. On the other hand, we realized after some initial tests that for our particular use case tilting the stereo cameras $25\degree$ towards the ground provided a better experience and ergonomics for the user. We designed a custom 3D printed attachment to fix the stereo camera to the VR device with the described offset as in Fig. \ref{fig:passthru}. 

\subsubsection{Video Capture}

Aligned with most of AR and VR applications, our solution is built using Unity 19.3. Consequently, we must be able to access the stereo camera feed from Unity to be able to process and render it within the immersive scenario. Our system has been designed to work both in tethered and wireless modes. Due to the device's particularities, both modes differ in the way that the video stream from the stereo camera is captured:

\begin{itemize}
    \item \textbf{Wireless Mode:} as the Quest 2 relies on Android as the base operative system, we built a native Android plugin to capture the video frames from the stereo camera. The plugin is based in the cross-platform UVC library \cite{libuvc} for webcam video capturing. We implemented a custom interfacing plugin to allow Unity to access the captured frames with as low latency as possible.  
    \item \textbf{Tethered Mode:} in this case, we capture the video frames as a simple webcam using Unity's C$\#$ API.
\end{itemize}

\subsubsection{Rendering Planes Placement}

\begin{figure}
  \centering
   \includegraphics[width=\linewidth]{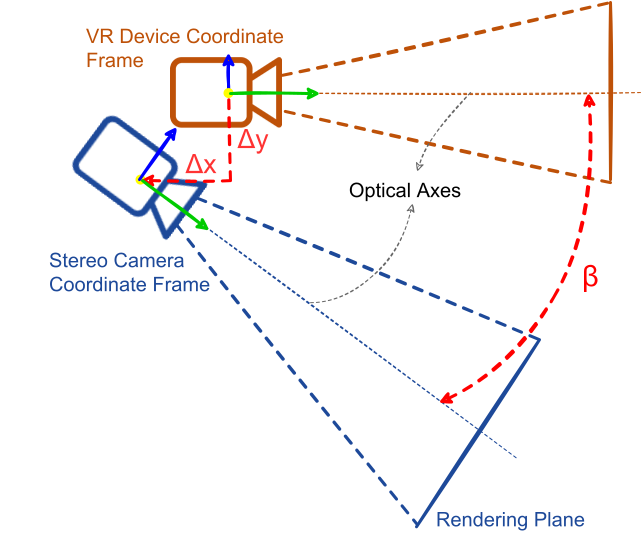}
  \caption{Schematized representation of the VR device's and stereo camera's rendering frame's coordinate frames. In orange, the VR device rendering camera coordinate frame. In blue, the rendering planes coordinate frame. In red, the parameters to be estimated using our custom calibration method. }
  \label{fig:Cameras_align}
\end{figure}

In any video pass-through MR application, the rendered stereo camera's feed and user's head movement must be perfectly aligned to avoid any user discomfort or VR sickness. This misalignment can come from three non-exclusive sources: lens distortion effects, inaccurate placement of the planes where the camera feed is rendered, and motion-to-photon delays. We need to apply the necessary calibration steps to reduce the effect of the aforementioned misalignment sources. 

\label{section:cameracalib}
The captured video frames from the stereo camera are rendered into two virtual planes (in blue in Fig. \ref{fig:Cameras_align}), which we refer to as rendering planes, one for each eye. The first step is to obtain the intrinsic parameters $[f_x, f_y, c_x, c_y]$ and distortion coefficients $[k_1, k_2, k_3, p_1, p_2]$ from the stereo camera so we can accurately correct the camera's lens distortion and properly scale the rendering planes. We used a well-known camera calibration technique described by Z. Zhang  \cite{CameraCalibration}. 

We used a custom shader for Unity to correct the camera's lens distortion using the estimated parameters, applied to the rendering planes' material. The shader implements the following distortion correction equation \cite{DistortionEquations} for each pixel $p_{xy}$:
\begin{align}
    x' = & x + \bar{x}(k_{1}r^2 +  k_{2}r^4 +  k_{3}r^6) + p_1(r^2 + 2\bar{x}^2) + 2p_{2}\bar{x}\bar{y} 
    \\
    y' = & y + \bar{y}(k_{1}r^2 +  k_{2}r^4 +  k_{3}r^6) + p_2(r^2 + 2\bar{y}^2) + 2p_{1}\bar{x}\bar{y},
\end{align} 
where $\bar{x} = x - c_x$, $\bar{y} = y - c_y$, $r = \sqrt{\bar{x}^2 + \bar{y}^2}$. Finally, we need to properly scale the rendering planes according to the intrinsic camera parameters, so that the real objects captured by the stereo camera are perceived with the appropriate size. Using an arbitrary distance($d_r$) between the rendering frames and the origin of the virtual cameras assigned to each eye (eye cameras) we can estimate the scale $S$ for a given resolution $[R_x,R_y]$: 
    \begin{align}
        S = [s_x, s_y] = [d_r\frac{R_y}{f_y} , s_y\frac{R_x}{R_y}].
    \end{align}

Once the camera has been correctly calibrated, we can solve the inaccurate placement of the rendering planes. The goal is to estimate the pose of the stereo camera with respect to the VR device's coordinate frame, as in Fig. \ref{fig:Cameras_align}. Specifically, we need to estimate the accurate position and orientation of the rendering frames with respect to the virtual cameras coordinate frames. The goal is to calculate the geometrical relationship $[\Delta{x}, \Delta{y}, \beta]$, depicted in Fig. \ref{fig:Cameras_align} between both coordinate frames.

While this calibration is traditionally achieved using a classic camera-IMU (inertial measurement unit) calibration method \cite{CameraIMU}, commercial devices don't provide access to raw sensor data with accurate time-stamps. Consequently, we decided to design our own workaround calibration method. Our method relies on the complementary hand controllers commonly available with current state of the art VR devices. These controllers are accurately tracked by the VR device relative to its own coordinate frame. Consequently, we designed and 3D printed an attachment for the Meta Quest 2 controllers which includes an Aruco \cite{Aruco} cube as the one shown in Fig. \ref{fig:passthru}. 

\begin{figure*}[!ht]
  \centering
  \includegraphics[width=1.0\linewidth]{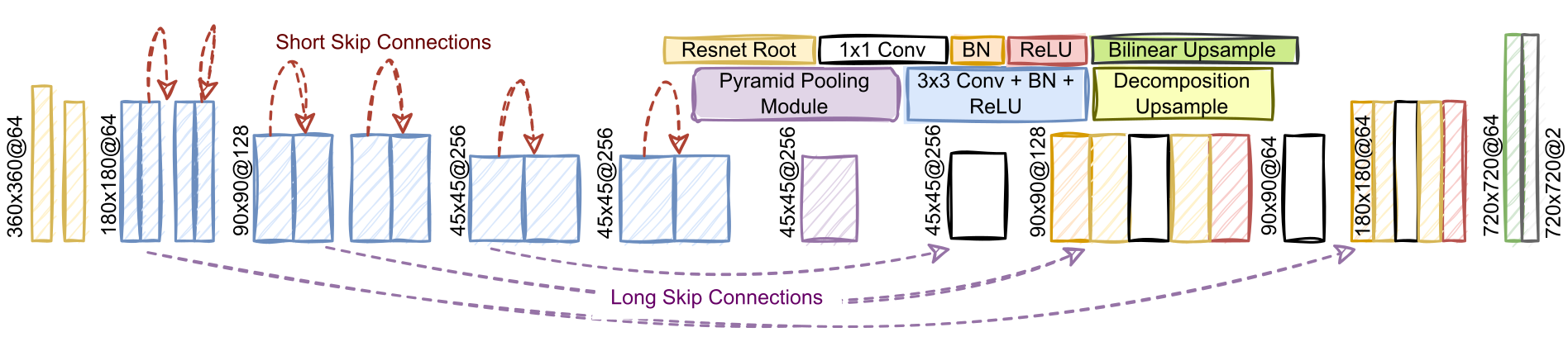}
  \caption{Deep learning architecture implemented and used for the task of egocentric human body segmentation, as described in \cite{gonzalez2022_realtimeseg}}
  \label{fig:thundernet}
\end{figure*}

Using a standard Aruco tracking library \cite{Aruco} written in C++ along with the estimated camera intrinsic parameters and distortion coefficients, we can accurately track the Aruco cube and, consequently, the correspondent controller's pose using the incorporated stereo camera. This estimation can be used to directly infer the geometrical relationship between the stereo camera's and VR device's coordinate frames. Consequently, we can directly estimate $[\Delta{x}, \Delta{y}, \beta]$, by estimating the required translation and rotation of the stereo camera's coordinate frame within the VR scene to perfectly align the VR controller pose as measured by the VR device and by the stereo camera. This relationship is constant, as the stereo camera is rigidly attached to the VR device. 

\subsubsection{Delay Alignment Buffers}
\label{sec:alignmentbuffer}
Even though the stereo camera lens distortion has been properly corrected and the video feed rendered on well-aligned and scaled rendering planes within the virtual scene, we need to introduce a method to remove the misalignment produced by the motion to photon latency. This latency produces an unnatural decoupling between the stereo feed, the head movement, and, consequently, the virtual scene. This artifact not only degrades the experience but can produce VR sickness. 

In a delay free system, we could just place the stereo camera coordinate frame in a fixed relative position and orientation with respect to the VR device coordinate frame. However, in a realistic scenario, this setup would produce the effect that the visual feed is delayed with respect to the user's head movement. To overcome this issue we add a delay alignment buffer, which stores the position and orientation corresponding to the stereo camera's coordinate frame. The size of the buffer corresponds to an arbitrary time $t_c$ which depends on the empirically estimated motion to photon delay. We assume $t_c$ to be constant for the same stereo camera and VR device models. Consequently, by placing and rotating the stereo camera feed always to the first pose stored in the buffer, we achieve an accurate head movement to stereo feed alignment.

\subsection{Egocentric Segmentation}
\label{sec:ego_segmentation}

This is a crucial step in our setup: we need an accurate real-time egocentric body segmentation algorithm which allows to identify which pixels, from the stereo camera feed, correspond to the user's body. Our algorithm is based on deep learning techniques. Particularly, it is based on the use of semantic segmentation networks \cite{guo2018review}. In this case, the segmentation is performed from first point of view (egocentric vision). 

The designed algorithm must meet two requirements: $i)$ real-time performance achieving an update rate above 60Hz, and $ii)$ high quality segmentation in uncontrolled conditions. Concerning the first requirement, we designed our architecture inspired in Thundernet \cite{xiang2019thundernet}, a very shallow semantic segmentation algorithm composed of: $i)$ an encoding subnetwork; $ii)$ a pyramid pooling module (PPM) that enables to extract features at different scales; and $iii)$ a decoding subnetwork (see Fig.\ref{fig:thundernet}). Several modifications were applied to the baseline architecture: $i)$ larger pooling
factors of 6, 12, 18, 24 were used in the PPM module to better extract features of larger input images, $ii)$ three additional long skip connections between encoding and decoding subnetworks for refining object boundaries were added, and $iii)$ weighted cross entropy loss function was used to handle class imbalance between human body (foreground) and background. 

\begin{figure*}[t]
  \centering
   \includegraphics[width=1.0\linewidth]{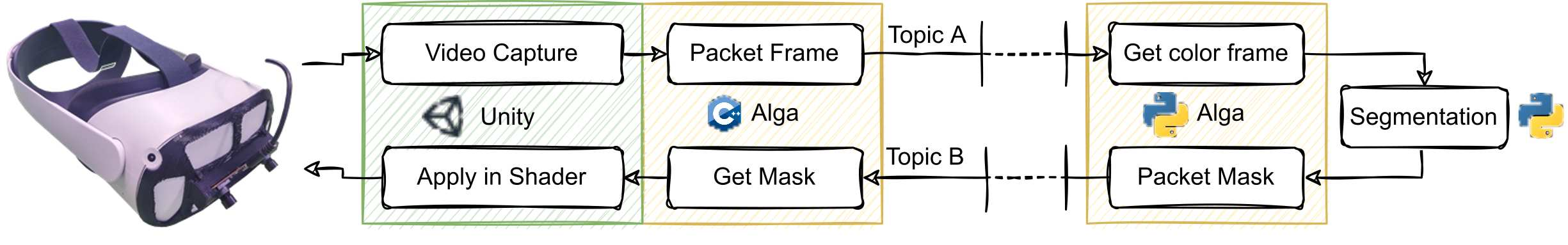}
  \caption{Final E2E simplified architecture, including the languages in which each agent was built. }
  \label{fig:finale2e}
\end{figure*}

To achieve high quality segmentation, a data-centric approach was followed, putting a strong emphasis on the variability of the training data. We created a dataset of almost $10.000$ images composed of three datasets: $i)$ Ego Human dataset, a semi synthetic dataset composed of egocentric body parts (arms, lower limbs, torsos) merged with realistic backgrounds that facilitates the process of groundtruth generation; $ii)$ a subset of THU-READ dataset, originally created for action recognition, whose segmentation groundtruth was created manually using Amazon Mechanical Turk, and $iii)$ EgoOffices: an egocentric dataset that was manually captured using a portable custom. The dataset was captured by more than $25$ users and multiple realistic scenarios, such as different apartments or office rooms. As the hardware setup involves the use of a stereo camera for providing stereo vision, to the user, we re-trained Thundernet architecture with images composed of two copies of the same images resulting in a $640\times480$ image, to replicate stereo vision. For more details, please refer to \cite{gonzalez2022_realtimeseg}.

\subsection{ZMQ-Based Offloading Architecture}

Our egocentric body segmentation solution can't fulfil the real-time requirements using mobile hardware, such as smartphones or untethered VR devices. Consequently, we decided to wirelessly offload the segmentation algorithm to a nearby server. The offloading architecture must be specifically designed to allow real-time processing, providing the highest throughput and lowest latency possible. For this reason, we used ZeroMQ\footnote{https://zeromq.org/}, a TCP-based architecture which has shown outstanding performance in most throughput and latency benchmarks \cite{ZMQBench}. We have developed our own ZMQ-based offloading architecture, which is already validated in realistic MR offloading use cases \cite{imxoffloading}. The architecture has been specifically designed to ensure high reliability, throughput, and low latency data transmissions. 

The architecture has been designed following a publisher-subscriber scheme in which one or multiple nodes advertise services to which other nodes can publish or subscribe to. These nodes can be defined as individual agents within the architecture which can subscribe or publish to an available data channel or topic. The architecture includes two main components: alga and polyp. Polyp is a data re-router which allows multiple nodes to subscribe and publish to different services. 

Alga is the main component which oversees data exchange between the VR client and the nearby server, following a publisher-subscriber logic. It is the core communication custom library which allows the direct data exchange between nodes, efficiently handling the data reception and transmission between them. While UDP-based protocols optimize the required throughput and transmission latencies, we preferred to prioritize the overall reliability of the architecture by using TCP as the underlying protocol. Besides, TCP allows port binding which considerably simplifies the server-client implementation. 

The architecture implements several types of publishers/subscriber according to the kind of data to transmit. Each type used a specific sort of custom packets, each divided in three sub-packets: topic, header, and data. For our particular application, we use the Picture and Unsigned 8-bits Picture packets. The first one is specifically designed to transmit individual JPEG-coded color frames. In our case, this packet type is used to transmit the stereo camera information. The second type is designed to transmit 8-bit single channel frames, specifically designed for segmentation masks. Consequently, this packet type is used to transmit the resulting segmentation mask back to the VR client. 

The architecture is built both in C++ and Python. On the server side, we use the Python version of the architecture to align with the egocentric body segmentation algorithm. As the VR client is built in Unity, we created a C++ plugin, allowing to use non-blocking asynchronous transmission and reception. This is necessary to avoid extra latencies coming from Unity's update rate not being synchronized with the camera capture or segmentation algorithm. 

\subsection{End to End System}

The final E2E setup is depicted in Fig. \ref{fig:finale2e}. The principal data flow of the E2E architecture is the following: 
\begin{enumerate}
    \item \textbf{Stereo camera frame capture: } As we have already described, the VR client runs Unity and is in charge of obtaining individual frames from the stereo camera video feed. The individual frames are moved to the underlying C++ plugin running our offloading architecture.
    \item \textbf{Transmit stereo color frame to server: } The offloading architecture, via Alga, is in charge of packing each frame and transmit them via TCP through an arbitrary topic. The segmentation server, runs another instance of Alga and receives the frames through the same topic. 
    \item \textbf{Egocentric body segmentation:} The server then infers the segmentation mask for each received frames.
    \item  \textbf{Transmit result back to VR client: }The mask is transmitted back to the VR client via Alga, through another arbitrary topic.
    \item \textbf{Shader application on rendering frames: } Finally, the client applies the received masks to the custom shader attached to the rendering frames. The shader is in charge of rendering exclusively the stereo camera pixels corresponding to the user's body according to the received mask. 
\end{enumerate}

\subsubsection{Hardware Setup}     
\label{sec:hardware}

The egocentric segmentation server was running on a PC with an Intel Xeon ES-2620 V4 @ 2.1Ghz with 32 GB of RAM and powered with 2 GPU GTX-1080 Ti with 12GB RAM each. On the VR client side, we considered two modalities: 
\begin{itemize}
    \item Standalone Mode: The immersive scene is running directly on the Meta Quest 2 HMD, to which the stereo camera is directly connected. 
    \item Tethered Mode: The immersive scene runs on a workstation to which a Meta Quest 2 is connected via an USB-C cable using the "Link" mode. The stereo camera is connected to the workstation. In this case the client workstation includes an Intel Core i7-11800H and a Nvidia Geforce RTX 3070 GPU.  
\end{itemize}

We used Unity version $2019.3$ on the client side, where the stereo frames are captured at a resolution of $1280\text{x}480$. For the validation experiments we used a Netgear R6400 router, providing symmetric wireless 200 Mbps when a single user is connected \cite{imxoffloading}.

\subsection{Performance Evaluation}

The proposed offloading architecture has been thoroughly benchmarked in terms of throughput and latency in multiple relevant scenarios and wireless technologies\cite{imxoffloading}. We decided to extend the benchmark by testing the architecture on the final E2E system. To obtain the following results we used the exact hardware setup described in Section \ref{sec:hardware}.

\subsubsection{Offloading Architecture and Video Pass Through} 

We first aim to evaluate the performance of our offloading architecture alone in the following setups: 
\begin{itemize}
    \item Wireless Scenario: Meta Quest 2 in Standalone mode using Wifi. We want to understand how the implemented system behaves on a device with limited computing resources. 
    \item Tethered Scenario: Meta Quest 2 connected to a workstation. In this case the system has considerably higher computing resources. The workstation is connected to the server via an ethernet cable, with a 1 Gbps interface. 
\end{itemize}

\begin{table}[tb]
\centering
  \caption{Performance evaluation results for the tethered and wireless scenarios along with the server processing times. }
  \label{fig:resultsArchitecture}
  \begin{tabular}{cccc}
    \toprule
         &  Tethered & Wireless & Server \\
    \hline
    Mean & 7.37 ms & 10.39 ms & 16.7 ms  \\
    Variance  & 14.37 ms & 19.38 ms & 0.006 ms  \\
    95-th Perc. & 14.0 ms & 18.0 ms & 31.2 ms  \\
  \bottomrule
\end{tabular}
\end{table}

In this first round of experiments, we removed the segmentation algorithm from the server, directly transmitting back to the client a single channel 8-bit mask of the incoming frame. This mask is obtained using a chroma-key algorithm which adds a negligible overhead on the server side. For each scenario, we repeated a set of 4 experiments each transmitting 1000 frames. No other users are connected to the router. We store the times that each frame takes from the moment it is loaded in memory on the client side to when the final mask is received from the server on the client side. The initial results are shown in the first two columns of Table. \ref{fig:resultsArchitecture}. Our E2E setup provides mean round trip times of 7.37 and 10.30 milliseconds on tethered and wireless scenarios respectively, with confidence intervals below 20 ms in both cases. 

\subsubsection{Segmentation Server}

In this case, the goal is to measure the total time consumed by the server in performing all the necessary steps for a frame: reception, necessary transformations such as scaling, and inference. Similarly to the previous set of experiments, we carried out 4 experiments each processing 1000 frames. We obtained a mean time of 16.7 ms, as shown in Table \ref{fig:resultsArchitecture}, which guarantees an update rate of 60 Hz, complying with MR applications real-time requirements. 

\subsubsection{End to End System}

Finally, by aggregating the results obtained in the previous two experiments we obtained E2E mean round trip times (RTT) of 29.67 ms and 32.39 ms. This latency is low enough to allow the use of delay buffers to ensure a proper alignment of the color frame and masks, as described in Section \ref{section:delayBuffers}.

\subsection{Delay buffers}
\label{section:delayBuffers}

\begin{figure}[t]
  \includegraphics[width=1.0\linewidth]{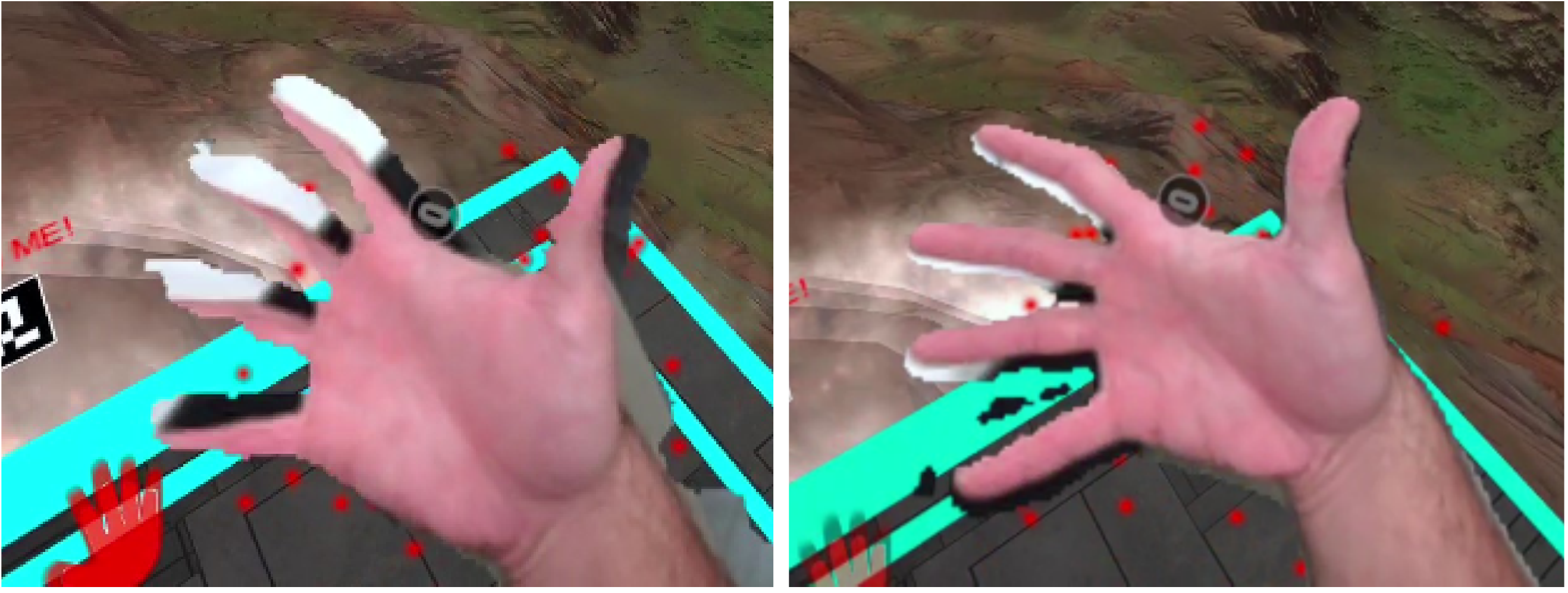}
  \caption{Left: Mask-color pixels alignment observed when no delay buffer is used. Right: alignment achieved using the delay buffer.\label{fig:handoff}}
\end{figure}

As the obtained RTTs values are higher than the device's update period of 16 ms, the hand pixels appear to be misaligned with respect to the segmentation mask. The misalignment comes from the fact that the segmentation mask arrives to the VR client an average of 32.29 ms later than its corresponding stereo pixels are loaded in memory and rendered. The feeling is as if the mask follows the actual hand. This artifact, shown in Fig. \ref{fig:handoff}, affects the overall experience and must me removed.

To overcome this issue, we decided to add a frame buffer of an arbitrary size according to the update rate of the camera and the measured mean RTT (32.29 ms). By doing this we overcome the artifacts generated by the system's added delays, ensuring the mask and its corresponding stereo pixels to be rendered simultaneously. Notice that the size in time of this delay buffer must be added to the alignment buffer described in Section \ref{sec:alignmentbuffer}.

\begin{figure*}[th]
  \centering
  \includegraphics[width=0.8\linewidth]{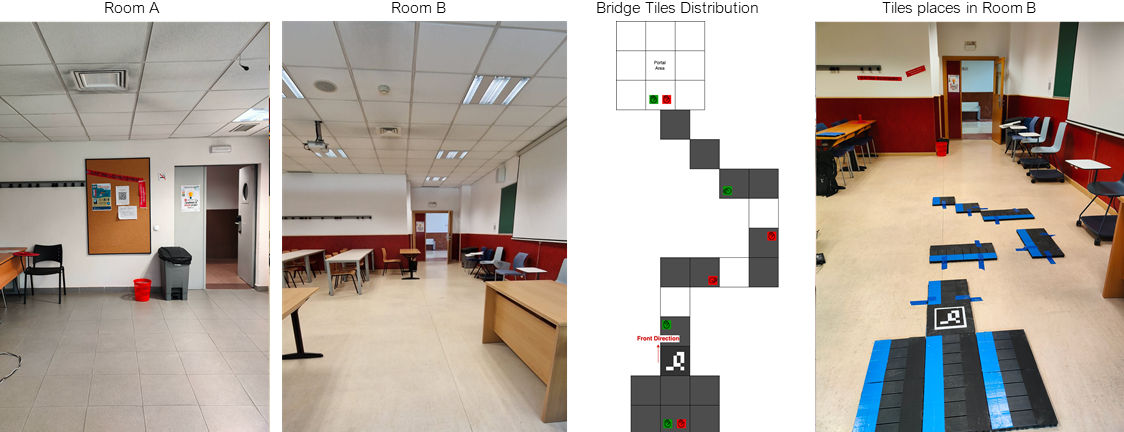}
  \caption{Left: the physical rooms used for the experiments. Right: the distribution of the tiles within the volcano path.}
  \label{fig:steps_rooms}
\end{figure*}

\begin{figure*}[th]
  \centering
  \includegraphics[width=1.0\linewidth]{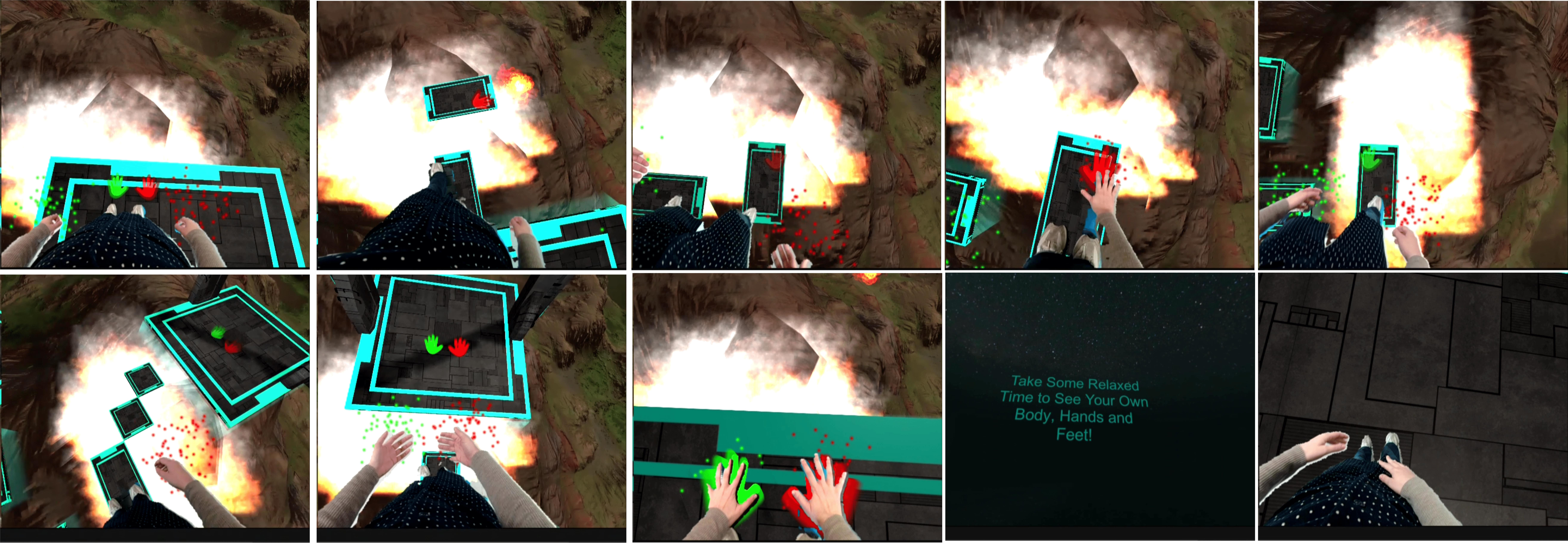}
  \caption{Detail of the different sequences happening in the immersive experience. Tiles are placed following the design depicted in Fig.\ref{fig:steps_rooms}}
  \label{fig:room}
\end{figure*}

\section{Subjective Evaluation}
\label{sec:immersive_experience}

Our study aims to investigate the role of full body video-based self-avatars in Mixed Reality and how our E2E implementation affects the overall experience in terms of presence, embodiment, visual quality, and acceptability.  
We compare our video-based avatar approach with a state-of-the-art hand tracking solution , as well as with our previous implementation of video-based avatars using chroma-key segmentation \cite{immersirve_gastronomic2019}.

\subsection{Research Questions}
\label{sec:rq}

The following research questions were established:

\begin{itemize}

\item \textit{RQ1} The inclusion of a stereo camera that allows video pass-through increases sense of presence and embodiment with respect to a virtual representation of your hands. Seeing your full body, including your feet, is relevant for the experience.

\item \textit{RQ2:} Deep learning based segmentation is a better solution for self-avatars in uncontrolled conditions than color-based segmentation.

\end{itemize}

\subsection{Immersive Experience: Design and Considerations}

We decided to create an immersive scene which forces the users to be aware of and use their entire body, involving demanding visuomotor tasks. To this aim, a volcano immersive environment was created. Users were told to walk over a set of narrow and non-contiguous tiles creating a path along the crater of the volcano, forcing the users to pay attention to their lower limbs while walking. To provide passive haptics, to increase the feeling of being there, we incorporate $1cm$ height steps, as can be seen in Fig.\ref{fig:steps_rooms}. The main purpose of the steps is to elicit visuotactile stimuli. To align the real and virtual steps we implemented a C++ plugin running an standard Aruco tracking library \cite{Aruco}, and the camera calibration parameters estimated in Section \ref{section:cameracalib}. We placed an Aruco marker as shown in Fig. \ref{fig:steps_rooms}. 

The whole path of tiles is not visible to the user from the beginning. On the contrary, the tiles are activated, in order, when users match their hands with neon-light representation of the hands shown on the floor\footnote{Hand tracking was provided by Meta Quest 2}. The neon-light hands are placed on the ground/tiles level so that the users need to touch with their own hands the actual surface, increasing the haptic sensations. The hands are tracked using Meta Quest 2 hand-tracking solution. Finally, at the end of the tiles path, we created a virtual portal that, once reached by the users, would teleport them to a friendly scenario. The goal of this final friendly scene is to allow the users to explore the virtual environment as well as its own body representation without any stress or pressure. The representation of the entire setup and the described details are shown in Fig. \ref{fig:room}.

\begin{figure*}
  \centering
  \includegraphics[width=1.0\linewidth]{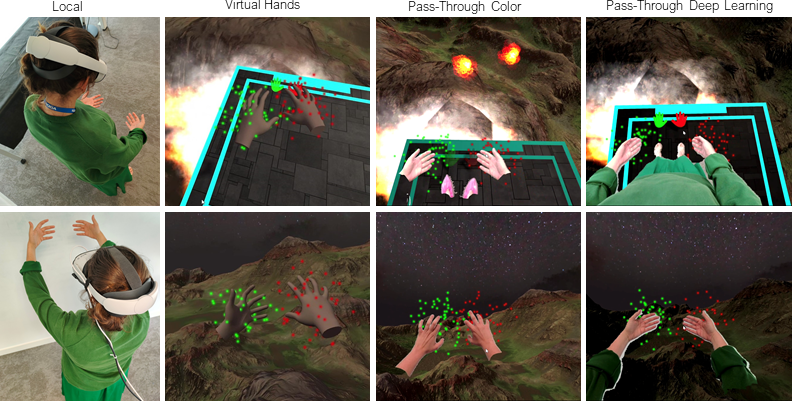}
  \caption{Conditions explored in the user study. From left to right: Local scenario, Condition 1 - Virtual hands, Condition 2 Pass-Through Color, and Condition 3 Pass-Through Deep Learning.}
  \label{fig:conditions}
\end{figure*}

\begin{figure}[h]
  \centering
  \includegraphics[width=1.0\linewidth]{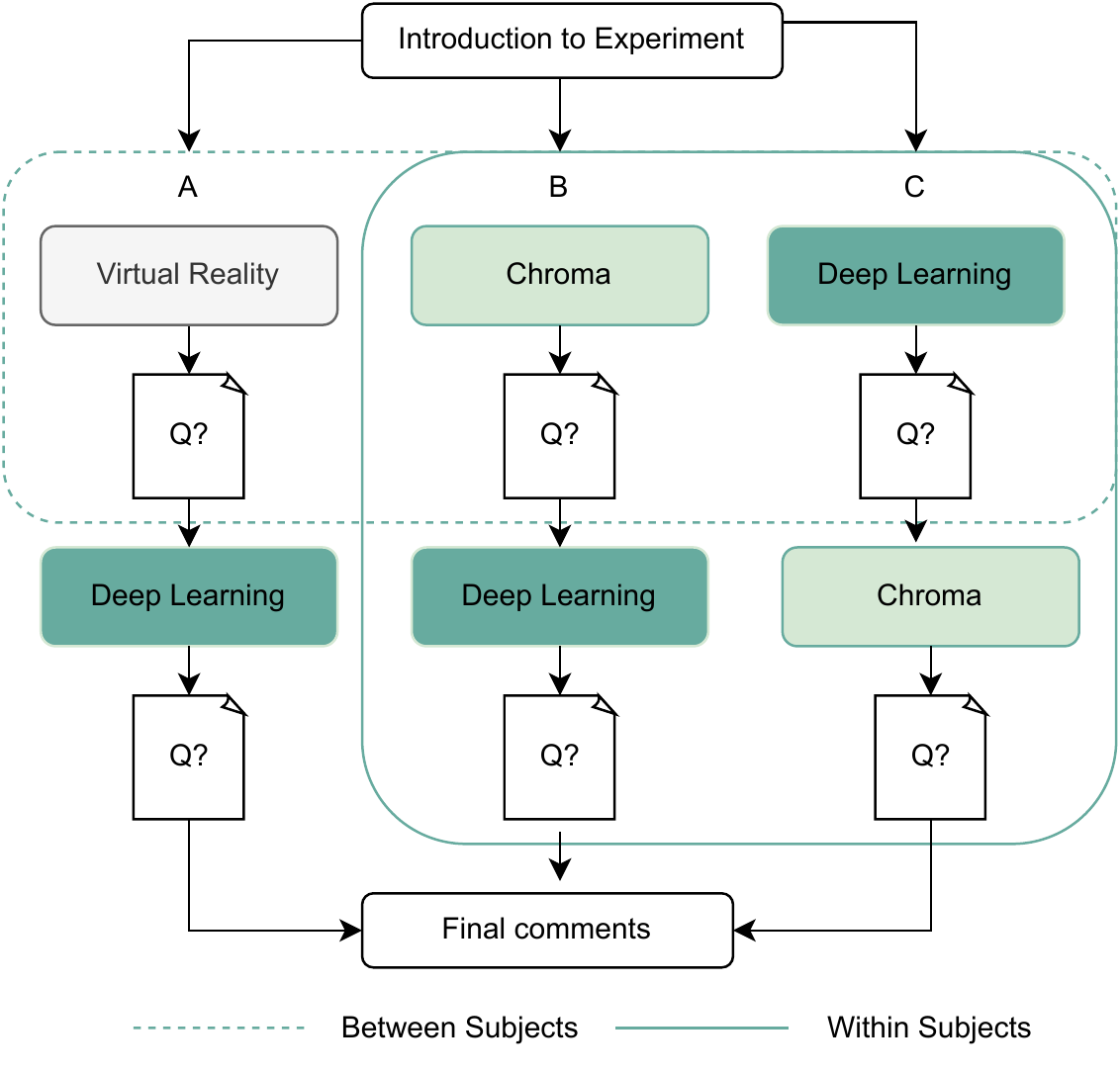}
  \caption{Diagram of the protocol followed to conduct between-subjects and within-subjects experiments}
  \label{fig:protocol}
\end{figure}

The total area occupied by the setup is 4m x 6m, using 16 steps in total. The used distribution is shown in Fig. \ref{fig:steps_rooms}. In Fig. \ref{fig:resultsArchitecture} we can observe RTT difference between the wireless and tethered setup to be sufficiently low to be negligible. Consequently, we decided to use the tethered setup in which the Meta Quest 2 is connected to a workstation running the scene to reduce the battery consumption and increase length of the experiment runs. 

\subsection{Conditions}

The main goal of the experiment is to evaluate the different modalities available to create the self-avatar user representations (see Fig.\ref{fig:conditions}). The following three conditions were explored:

\begin{itemize}
    \item \textbf{Virtual Reality:} the user is represented with the virtual hand models provided by the Meta Quest 2. 
    \item \textbf{Chroma:} users are represented as the output of a simple algorithm that masks everything in the range of skin colour, similarly as in \cite{immersirve_gastronomic2019}. To ensure that users could see their feet as a reference to walk, we asked them to wear some pink booties, which were perfectly detected by the color-based filter, as can be seen in Fig.\ref{fig:conditions}.
    \item \textbf{Deep learning}. the user is represented by the output of our semantic segmentation algorithm reported in Section \ref{sec:ego_segmentation} using the described E2E architecture.
\end{itemize}

\subsection{Protocol}
At the beginning, participants were briefed about the experiment and the purpose of it, and asked to sign a consent form. Three modalities were set up: A, B, and C. To counterbalance the order effect, users were evenly distributed among modalities. On each modality, users need to perform the experiment under two conditions, as depicted in Fig.\ref{fig:protocol}. After each condition users answered the questionnaire described in Section \ref{sec:questionnaire} .

Experiments took place in Madrid during three consecutive days in March 2022 in a local university. The first and third day room A was used, while the second day room B was used (see Fig.\ref{fig:steps_rooms}). 

It is worth noting that, for safety reasons, users under modality A did not use the 1 cm steps described above, since the Virtual Reality condition did not allow them to see their feet.

\subsection{Questionnaire}
\label{sec:questionnaire}

After, providing some demographics information about age, gender, and previous experience with VR, users were asked to fill a $30-item$ subjective questionnaire, where $12$ were used to measure embodiment, $7$ for presence, $6$ for visual quality, and $4$ for acceptability. 

\textbf{Embodiment}. We have used the embodiment questionnaire (EQ) proposed by Gonzalez-Franco and Peck \cite{gonzalez2018avatar}. We chose to include a subset of $13$ items, namely: Q1, Q2, Q3 for ownership, Q6, Q7, Q8 and Q9 for agency, Q14 for location; Q17, Q19, and Q20 for external appearance, and Q25 for external stimuli (modifying it to \textit{"I had the feeling that if I fell off the bridge I was going to hurt myself"}). Unlike in the original questionnaire, Q20 (``I felt like I was wearing different clothes from when I came to the laboratory'') have been reversed, since now the users should be able to see their own clothes.
Items from tactile sensation factor were not included as they were not addressed in our study. A total score has also been computed following \cite{gonzalez2018avatar}, as the weighted average of all items, where items Q1-Q14, which belong to the main embodiment components (ownership, agency, and location) have double weight than the rest of the items (Q17-Q25).

\textbf{Presence}. We have used Witmer and Singer's presence questionnaire version 3 (PQv3)  \cite{witmer2005factor}. PQv3 contains $32$ items to explain sense of presence using the four different sub-dimensions: involvement, interface quality, adaptation / immersion, and sensory fidelity. We use a subsampling with $7$ items, as it shows a good correlation with the full questionnaire in the overall score (computed as the mean of all items), as well as in the represented sub-dimensions: involved, adaptation / immersion and haptic sensory fidelity \cite{perez2021ecological}.



\textbf{Visual Quality}. To complete the evaluation of the experience, we have included additional questions to assess the subjective perception of the visual quality of the different elements \cite{immersirve_gastronomic2019}.
The visual quality of the virtual \textit{Environment}, the virtual \textit{objects}, and user's own \textit{Body}, \textit{Arms}, and \textit{Legs} were assessed using ITU-T P.913 Absolute Category Rating scale (ACR).
Since some of the elements may not appear under some conditions, an additional category was included indicating that the element was not visible at all, which was rated as 0.
Additionally, the annoyance of elements from the local \textit{Background} that were misclassified as foreground by the segmentation algorithm (false positives) were rated using ITU-T P.913 Degradation Category Rating (DCR).
A Total score has also been computed using the average of all scores (including both ACR and DCR ratings).

\textbf{Acceptability}. Four additional questions were included to explore people acceptability. Cybersickness was assessed using the single-item Vertigo Score Rating (VSR) scale \cite{perez2018towards}, as recommended by ITU-T P.919.
We use the ratio of users with severe symptoms ($VSR \le 2$) as evaluation score.
Global Quality of Experience (QoE) was asked using ITU-T ACR scale, and then the proportion of good-or-better (GoB) ratings was extracted \cite{hossfeld2016qoe}.
Finally, the Net Promoter Score was extracted with two computations: the traditional one, based on the self-reported probability of recommending the game to a friend (“NPS-R”) \cite{reichheld2003one}, and a variant based on 
the self-reported willingness to pay (“NPS-P”) \cite{villegas2020realistic}.

 
\textbf{Qualitative evaluation}. Users had two moments to provide free-form feedback, one answering \textit{why would you recommend the experience}, and a final comment free-form text about the overall experience. 



\subsection{Statistical analysis}
The design of the experiment allows for two types of analysis (see fig \ref{fig:protocol}): a between-subject analysis between the first execution of the three conditions, aimed at comparing the Virtual Reality baseline solution with both video-based avatars (\textit{RQ1}), and a within-subject analysis between Chroma and Deep Learning implementations, to address \textit{RQ2}.

Presence, Embodiment, and Video Quality measures can be modeled as normal variables, and therefore we will analyze them using parametric statistics \cite{narwaria2018data}. One-way ANOVA will be used for the between-subject analysis, with Bonferroni-corrected pairwise Student T-tests as post-hoc analysis. One-way ANOVA with repeated measures will be used for the between-subject analysis. No post-hoc test is needed, as there are only two conditions.

Acceptability scores do not use a (weighted) average of the different ratings to provide a final score, but an analysis of the distribution of the ratings.
Consequently, to test for the significance of the differences of the acceptability measures, Mann-Whitney U-tests will be applied to the distributions of ratings, with Bonferroni correction for the within-subject analysis.

A significance level of $p < 0.05$ is established for all the tests.

\begin{table}[tb]
\centering
  \caption{Distribution of users for Modalities}
  \label{tab:distribution}
  \begin{tabular}{ccccc}
    \toprule
        \textbf{Modality} &  A & B & C & Total \\
    \hline
    Number of Men & $11$  & $10$  & $7$  &  $28$ \\
    Number of Women & $7$  &  $10$  & $13$ & $30$  \\
    Total & $18$ & $20$ & $20$ & $58$ \\
  \bottomrule
\end{tabular}
\end{table}

\begin{figure*}[p]
    \centering
    
    \begin{tabular}{cc}
        \includegraphics[height=0.6cm]{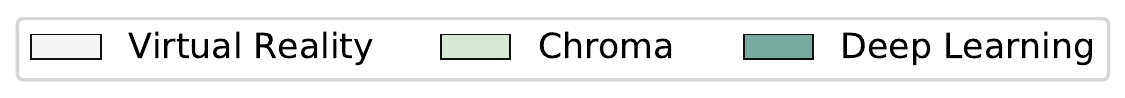} & 
        \includegraphics[height=0.6cm]{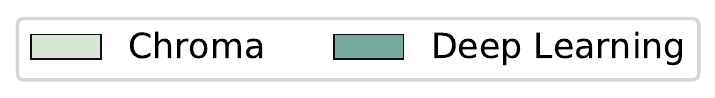} \\
        \includegraphics[width=0.9\columnwidth]{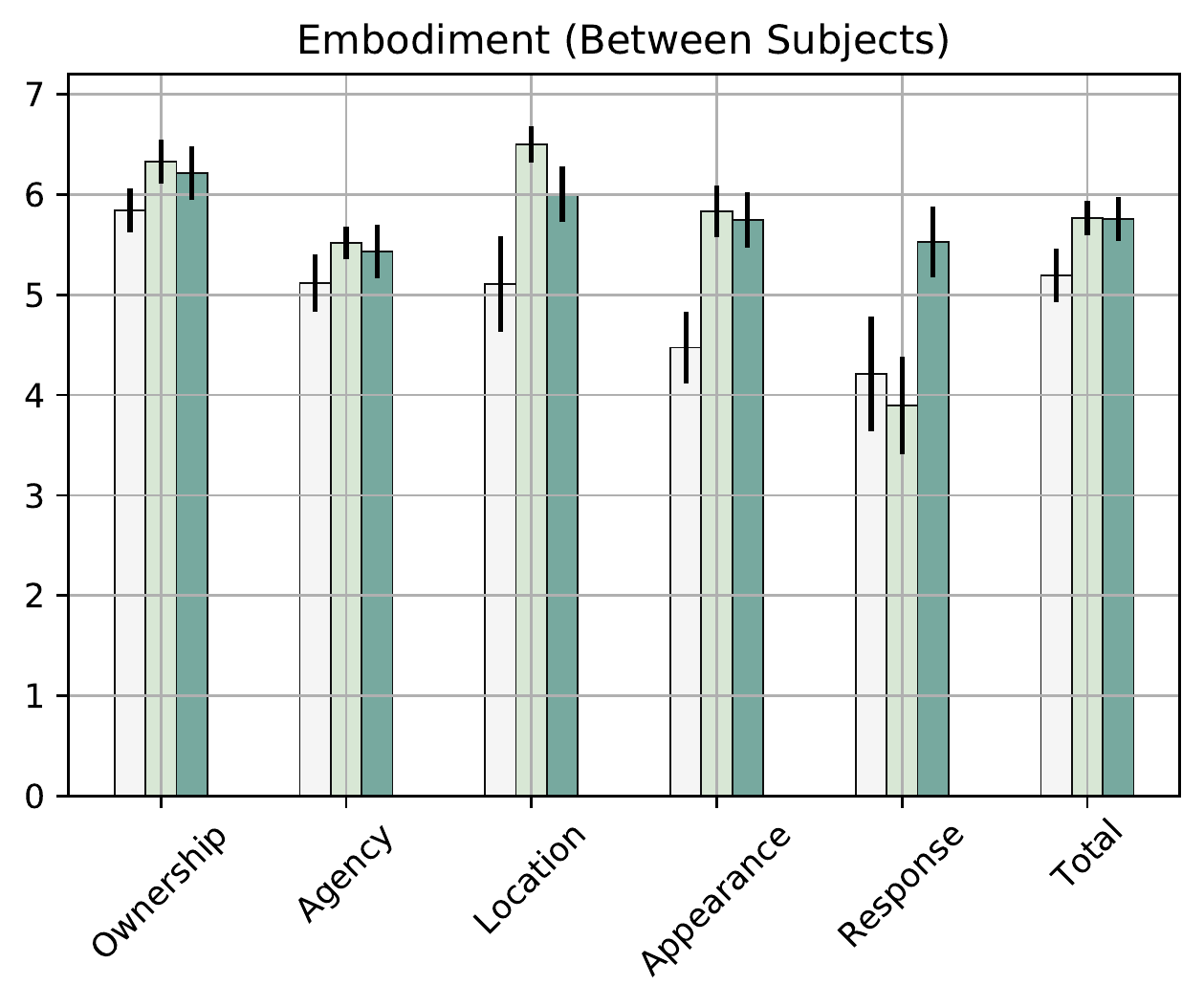} &
        \includegraphics[width=0.9\columnwidth]{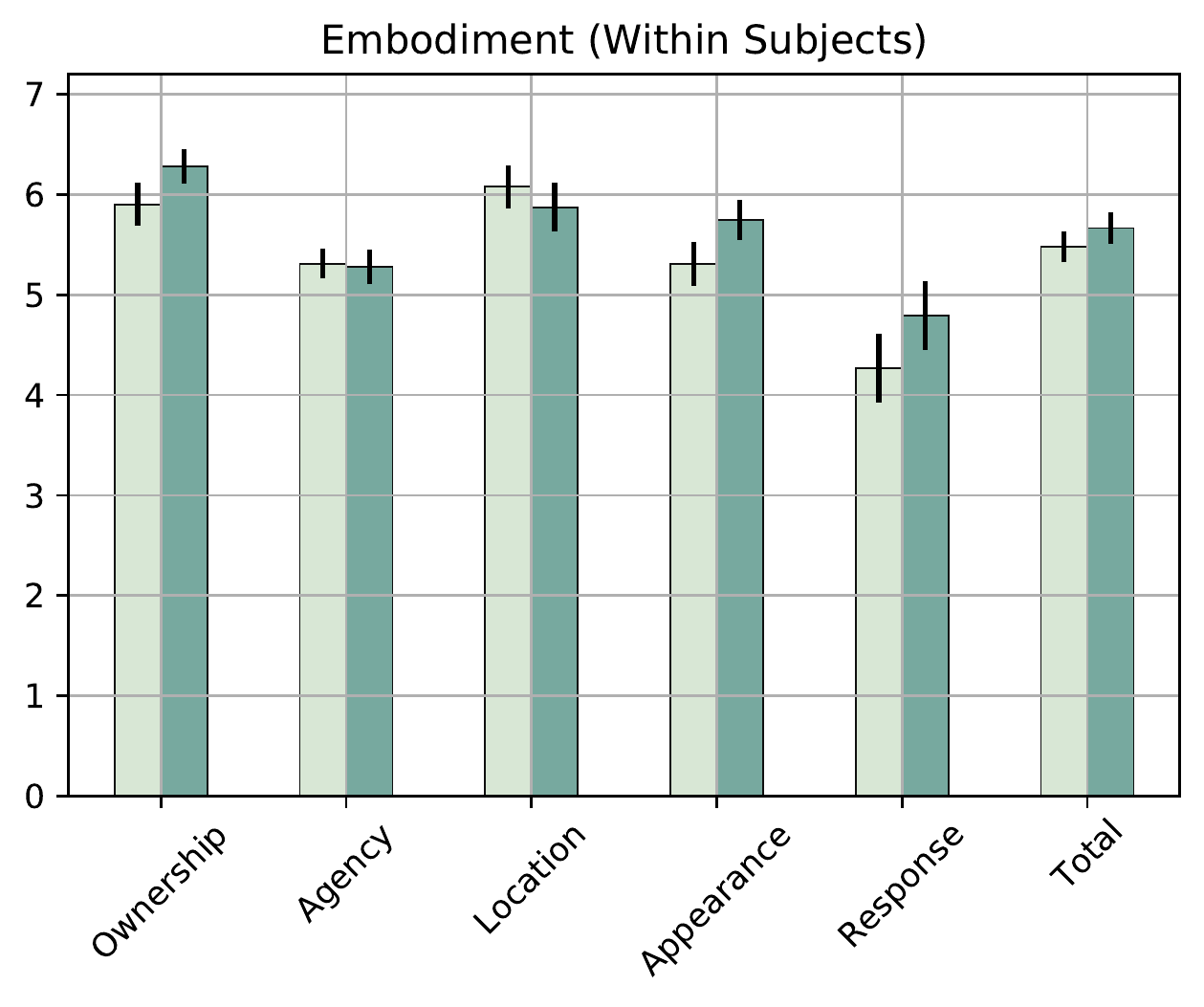} \\
        \includegraphics[width=0.9\columnwidth]{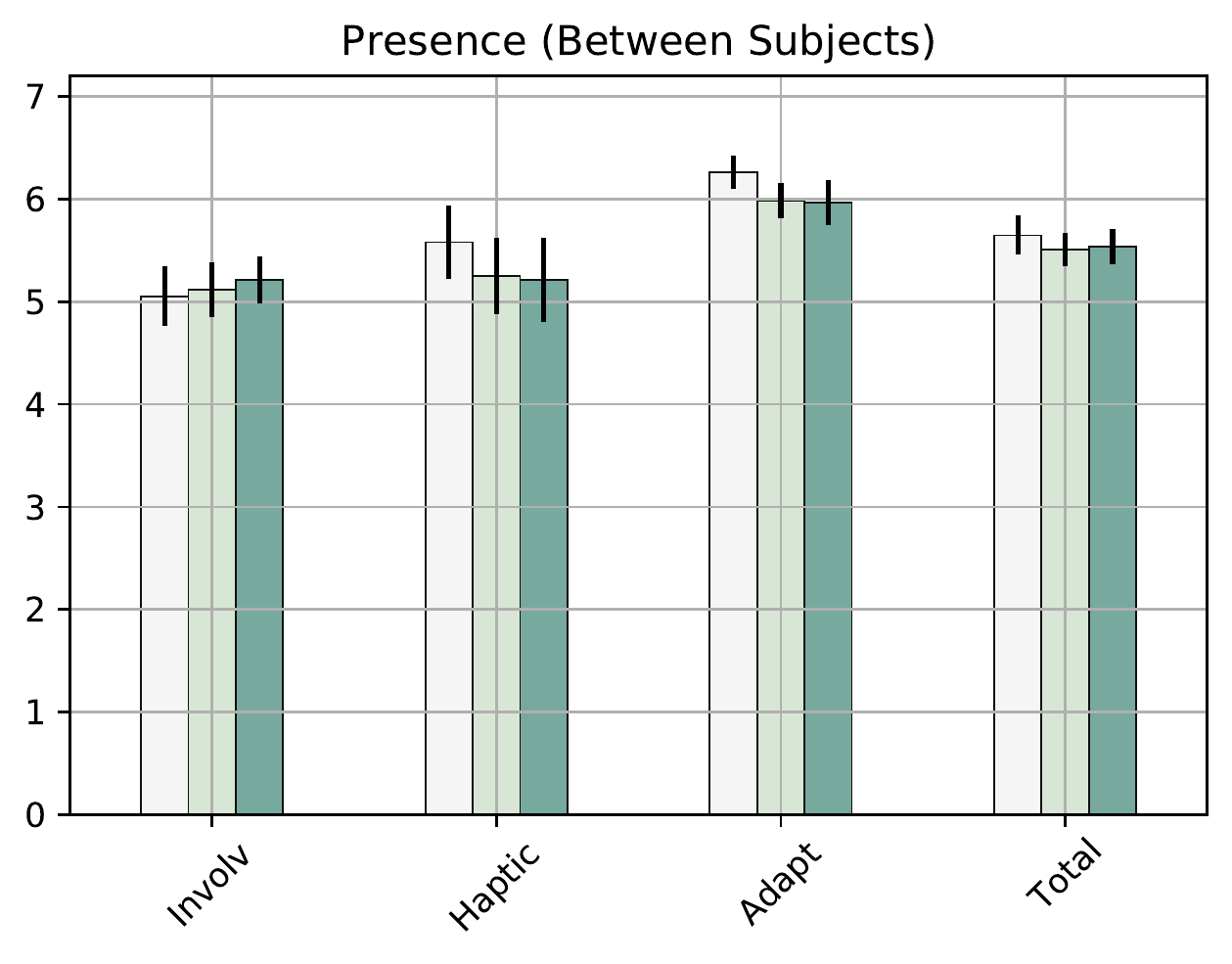} &
        \includegraphics[width=0.9\columnwidth]{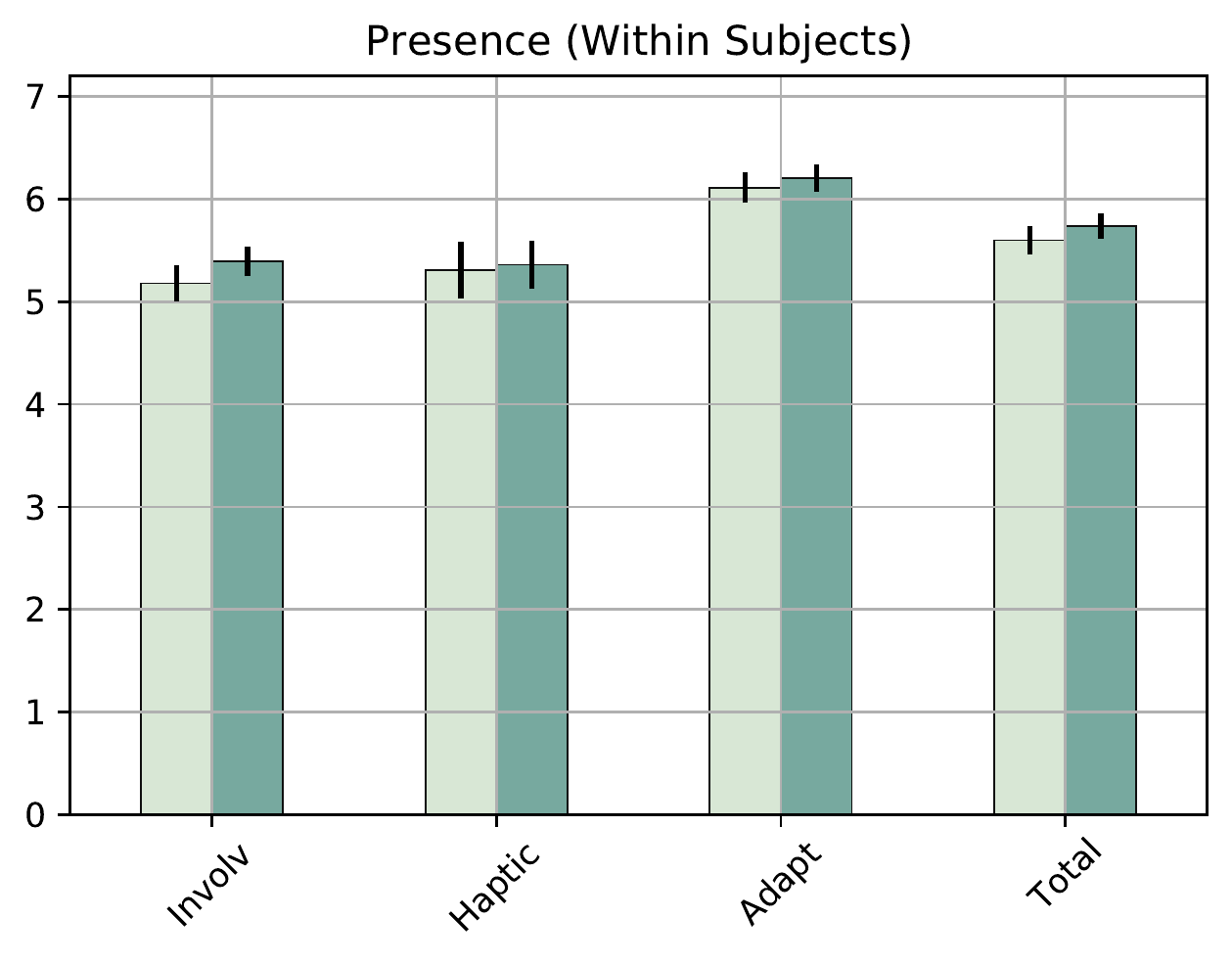} \\
        \includegraphics[width=0.9\columnwidth]{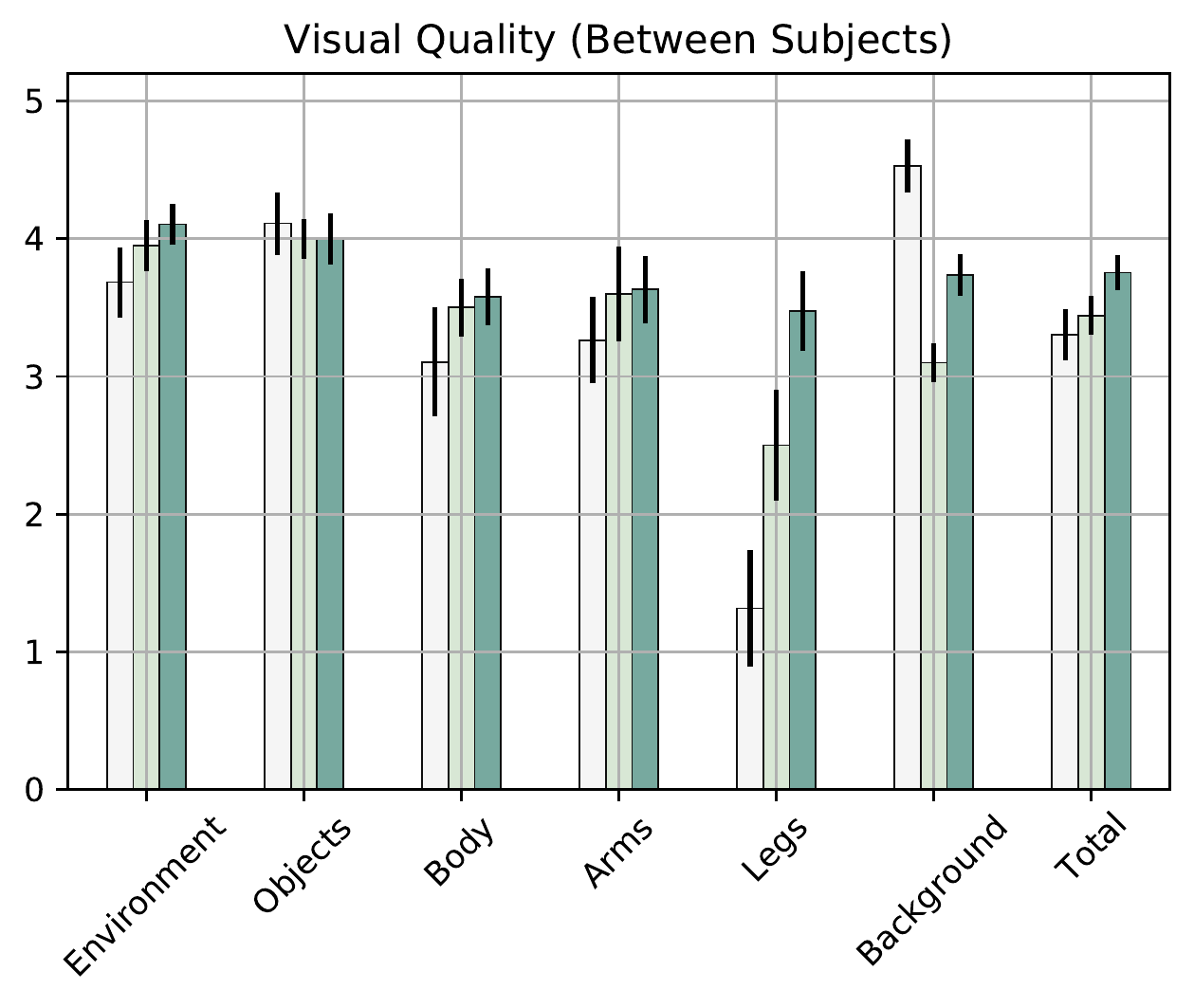} &
        \includegraphics[width=0.9\columnwidth]{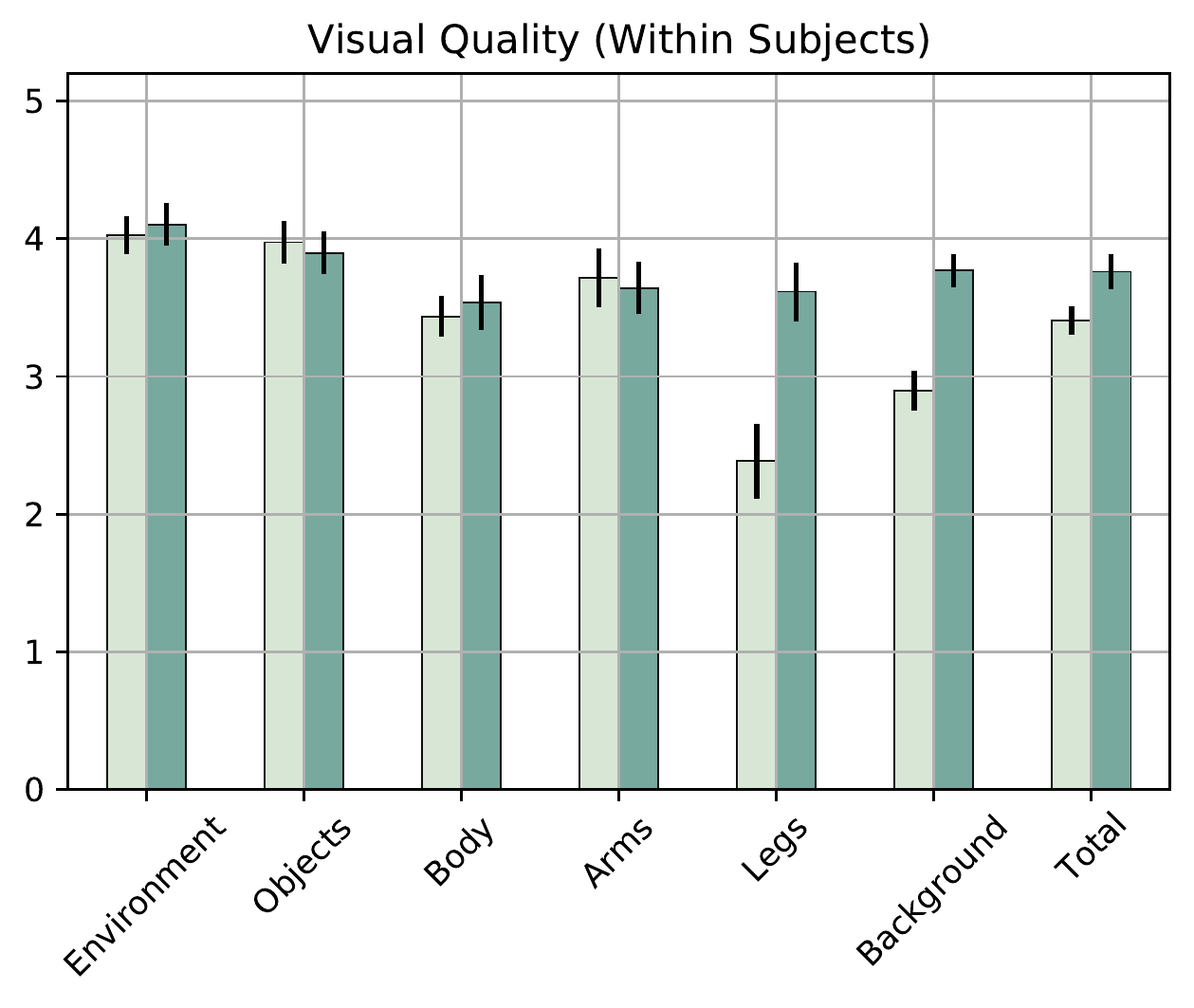} \\
    \end{tabular}
    
    \caption{Results for Embodiment (top), Presence (middle), and Visual Quality (bottom) questionnaires, both for Between Subjects (left) and Within Subject (right) analysis. Bars show Mean values for each of the measures, and error bars show the Standard Error of the Mean (SEM).}
    \label{fig:results}
\end{figure*}

\subsection{Participants}

The game was evaluated by $58$ voluntaries ($30$ female and $28$ male. Participant ages ranges from $18$ to $63$ (M=$22.16$, SD=$8.62$), most of them students from a local university and some of them staff, with no implication in the development of the project at all. All users were Caucasian but one woman, who has Black ethnicity. Table \ref{tab:distribution} shows a fair homogeneous distribution between modalities and gender distribution between modalities. $22$ of the $58$ perform the experiment in Room B (see Fig \ref{fig:room} right)., while the remainder $36$ performed the experiment in Room A (see Fig \ref{fig:room} left). One women assigned to C modality could only accomplish the first condition, as she reported extreme fear of conducting the experience one more time.

\section{Results}
\label{sec:results}

Fig. \ref{fig:results} shows the main quantitative results for Embodiment, Presence, and Visual Quality.

\subsection{Embodiment}

The quantitative results of the different sub-scales of the Embodiment Questionnaire, as well as a total value aggregating all items, are shown on Fig. \ref{fig:results} top row, for both between and within subject analysis. Results are shown in 1-7 Likert scale. All scores are high, indicating that the sense of embodiment is high under all conditions.

Between-subject analysis shows a significant effect of the video-based algorithms in Location ($F_{2,55} = 4.56$, $p = .015$, $\eta^2 = .14$), Appearance ($F_{2,55} = 6.48$, $p = .003$, $\eta^2 = .19$), and Response ($F_{2,55} = 3.26$, $p = .046$, $\eta^2 = .11$) scales. Pairwise T-Tests show significant differences between Virtual Reality and both Chroma and Deep Learning for Appearance and Location; and between Chroma and Deep Learning for Response.

Within-subject analysis shows a significant effect between Chroma and Deep Learning algorithms in Ownership ($F_{1,38} = 4.37$, $p = .043$, $\eta^2 = .02$) and Appearance ($F_{1,38} = 6.13$, $p = .018$, $\eta^2 = .03$). 

In summary, we can see moderate improvements in some Embodiment components when replacing the conventional virtual hands with a full-body video-based avatar, especially in terms of Location and Appearance. We observe small but significant improvements when we compare both video-based avatar solutions.

\subsection{Presence}

The quantitative results of the different sub-scales of the Presence Questionnaire, as well as a total value aggregating all items, are shown on Fig~\ref{fig:results} middle row, for both between and within subject analysis. Results are shown in 1-7 Likert scale.

Presence mean values are high for all sub-scales: 5.1 for Involvement, 5.2 for Haptic sensory fidelity, and 6.1 for Adaptation. However, no significant differences are found between conditions in either between or within subject analysis.

\subsection{Visual Quality}
Visual Quality scores are shown on Fig. \ref{fig:results} bottom row, for both between and within subject analysis. Mean Opinion Score (MOS) values are shown in the 0-5 extended ACR scale, except for Background, which uses 1-5 DCR scale. A Total (average) score is also shown.

Environment and Objects are common for all the conditions, and the show a MOS value about 4.0 (\textit{Good}), without significant differences.
The perception of Body and Arms has a moderately lower MOS, between 3.0 (\textit{Fair}) and 3.5, and the differences between algorithms are not significant either.

There is a significant difference on the perception of the Legs. 
This is obvious for Virtual Reality, where no legs were shown in the virtual scene, but also between Chroma and Deep Learning algorithms, in both between and within subject comparisons. In the within-subject comparison, the MOS for Chroma and Deep Learning is 2.38 and 3.61 respectively, which means an increase of more than 1 quality level.

The same difference exists in the perception of the Background false positives. In this case, the highest quality is obtained in the Virtual Reality scene, where no false positives exist. The within-subject comparison between Chroma and Deep Learning shows a significant MOS values differences, with results of 2.90 and 3.77 respectively. 

Consequently, there is a significant difference in the Total score both for Between-Subject and Within-Subject comparisons.

\subsection{Acceptability}

\begin{table}[tb]
    \centering
    \caption{Acceptability results summary}
    \begin{tabular}{c|ccc|cc}
         & \multicolumn{3}{c|}{Between} & \multicolumn{2}{c}{Within} \\
         & VR & Chr & DL & Chr & DL \\
        \hline
        Cybersickness & 0 & 5\% (1) &  5\% (1) & 2.5\% (1) & 0 \\
        Good-or-Better & 79 \% & 95 \% & 90 \% & 87 \% & 92 \% \\ 
        NPS-R & 47 & 70 & 53 & 51 & 61 \\
        NPS-P & 68 & 45 & 37 & 44 & 44 \\
    \end{tabular}
    \label{tab:acceptability}
\end{table}

Table \ref{tab:acceptability} shows the acceptability results. Only two participants reported severe cybersickness effect, both in the first execution of video-based avatar conditions: one in Chroma, one in Deep Learning. The latter did not want to do the second condition, and therefore her results have been excluded from the within-subject analysis altogether.

Quality of Experience ratings were extremely high, with above $80\%$ of the participants reporting Good or Excellent QoE. Net Promoter Scores were also high, mostly in the range $40-70\%$. 

When comparing the acceptability of the three conditions, no strong differences can be observed, as different acceptability indicators show different results. Furthermore, the U-test analysis of the underlying distribution of the scores does not show any significant difference.Therefore we can conclude that all three conditions have similar (and high) levels of acceptability.





\subsection{Qualitative evaluation}

Several insights could be also inferred from analyzing free text comments given by users. People in general were grateful to experience this new technology, stating in most cases that it was a very interesting/incredible/pleasant experience, reporting to have fun while doing the experiment, and envision it as \textit{the future}.

In all conditions, most users reported to feel teleported to the place \textit{"I felt like I was really there"}, and also to experience the threaten of the experience: \textit{"It's interesting to see how your body reacts unconsciously even though you know nothing will happen to you"} or \textit{"Even though it is a game it gives you the feeling that you are going to fall down [...] partially tricks your brain [...] there was a moment when I had to hold on because I felt like I was really going to fall"}. People in general reported good perception of the distance, specially related to touching the tiles, but show some uncertainties with respect to their height perception. From the perspective of the game itself, people felt that the game it's too short, and that adding more functionalities could make the game more fun. Some people wearing prescription glasses, reported a more comfortable experience when they took them off. 

In the VR scenario, some users were very happy to see their own hands without using controllers, while at the same time, reported that it could be even better if more body parts were perceivable. Indeed, other reported \textit{"disconcerting experience of not knowing exactly where to walk"}. With respect to the perception of false positives, users perceive less of them in the deep learning condition, vs the chroma condition: \textit{"the false positives were very low compared to the previous experience and the avatar looked much better than in the previous experience, it was quite good"}, or this comment after conditions PT-CL: \textit{"the second time  I tried the game it looked worse, I saw many more parts of the class, which made me feel more out of the game environment"}.

In general, in all deep learning responses, people show high acceptance to see their own body, reporting that makes them feel more immersed, perceiving the experience as more realistic. Besides, user also appreciated to be able to see their accessories such as watches, jewelry, painted nails, etc. Specially in modalities A, and B,  users reported a high difference with respect to the previous conditions, where only certain body parts could be seen. For instance: \textit{"It has been really impressive to see the change and be able to see my own body including my clothes and many details. It honestly blew me away. I've never had a VR experience where I could actually see my body and everything"} or \textit{"It looks like you're the one going through the catwalks and, moreover, the avatar looks exactly like you"}. Moreover, some people reported the benefits of seeing your own legs: \textit{"Seeing my legs made me feel better in space"}.

In the chroma scenario, the feedback could be different depending on whether users where wearing clothes in the same range as the skin color \textit{"The avatar was very well done"}, or not \textit{"the credibility is less because the body is not completely represented (I could not see my legs)"}

Users also reported that there is still room for improvement concerning seeing your entire body with deep learning. One thing is the fact that segmentation boundaries are not perfect, and tends to be slightly bigger (thick edge) than the actual body: \textit{"The fact of showing one's own body helps a lot to the perception of space and movement of oneself, however the noise, that is produced around the body distracts me a lot from the scene, diverts the attention too much and took me out of the experience"} or \textit{"I think it is a unique experience, however I think the quality of the avatar image can be improved"}. In this case, users perceived that the segmentation performed with color-based is more precise, although can only segment skin color parts, e.g. \textit{"in this experience, the hands looked more realistic because the cropping was better, however I was not able to see my own clothes and I saw most of the furniture in the classroom"}. In general, tracking was perceived very stable in all conditions, although the presence of long hair can block it the device's sensors.

\section{Discussion}
\label{sec:discu}


\subsection{Implementation of a Video-based Self-Avatar System for MR}

We have presented our fully functional Video-based self-avatar E2E system for MR. Our implementation allows to run the proposed segmentation algorithm in real-time on a commercial VR device. We achieved E2E latencies below 11 ms if no algorithm is running on the server. Consequently, the system allows to use other computationally demanding solutions on standalone VR devices such as object tracking or camera-based full body tracking, extending the current boundaries of MR solutions and applications. 

The system still have some limitations and we had to assume a set of trade-offs in terms of segmentation accuracy, frame rate, resolution, etc. Consequently, the presented solution can not be fairly compared to much more mature solutions such as the virtual hands solution based in hand tracking provided by Meta or other vendors. However, we must remark that the presented technology is in the boundary of the state of the art which leaves room for improvement and it can be a starting point for future distributed and novel MR solutions. 

\subsection{Video-based Self-avatars Evaluation}

Regarding the first research question established in section \ref{sec:rq}, our qualitative analysis shows that video-based self-avatars provide marginal improvement only in some components of the sense of embodiment with respect to the Virtual Reality condition. No significant difference is observed in presence.

Regarding the second research question, the Deep Learning algorithm offers better visual quality than the Chroma in uncontrolled environments, both for the legs and for the background perception. These differences result in small improvements in Ownership and Appearance factors of the sense of embodiment. No significant difference is observed in presence either.

We can extract several interpretations from this. First, all the conditions show high levels of sense of presence and embodiment. Since in all cases users are immersed in a virtual environment with which they interact with their own hands (either ``real'' or virtual), the degree of embodiment and presence achieved by the baseline condition is high. Therefore any observed improvement must be necessarily small. Second, the maturity of the virtual reality technology is much higher than the video-based avatars and, consequently, the implementation of VR looks more like an actual finished product than the video avatars.
Somehow the benefits of seeing their own body may be limited by the limitations on the implementation (e.g false positives). For example, VR only shows the hand (compared to the full arms of other solutions), but it offers better resolution in the hand image. Both facts counterbalance and, as a result, the video quality for the Arms is similar in all conditions.

A similar situation happens between Chroma and Deep Learning: the former is only able to detect the hands and some colors of clothes (red or yellow); however, these elements are segmented with better pixel precision than the Deep Learning algorithm. Besides, differences regarding the amount of body parts segmented between Chroma and Deep Learning might be reduced when subject are wearing clothes with similar colors to the skin. This was the case for $18$ out of the $58$ considered subjects. 

These findings are also supported by qualitative evaluation given by users: they appreciate the possibility of seeing completely your own body and at the same time reported room for improvement. Finally, we also experienced that some people tend to forget whether they have seen their own body limbs or not, which increases some uncertainty to the results.


Finally, it is also possible that the main benefits of using video-based avatars cannot be measured in terms of presence or embodiment. Other aspects of the overall Quality of Experience should be explored.

\subsection{Evaluation methodology}

First of all, we are evaluating presence, embodiment and video quality using already existing questionnaires and methodologies, but none of them was designed for video-based avatars. As a consequence, there are some limitations.

Presence Questionnaire (PQ) was developed for VR, and Embodiment Questionnaire (EQ) was designed for computer-generated and animated avatars. When used in a different scenario, some items may not completely apply, or they could be even misunderstood by the users. For instance, we have seen that Q8 (\textit{I felt as if the movements of the virtual body were influencing my own movements}) shows a range of values completely different from other items, which might show that is not being well understood by the users. Other items such as Q2, Q15, Q17, or Q20 might create some confusion to the users due to the particular case of this technology showing the user's appearance as it is. 

We have also observed some limitations of some questions related to visual quality, as we are using a single question to capture at the same time quality perception related to \textit{how much are you seeing from your body} and \textit{how sharp, accurate and stable this vision is}. These two aspects of the quality should be probably rated separately. In what concerns NPS-P, people reported that they will be willing to pay more if the game itself were longer. 

\subsection{Comparison with previous works}

Previous works have also observed that a realistic virtual hand representation elicits stronger sense of ownership \cite{argelaguet2016role}. Later, Fribourg \textit{et al.}\cite{fribourg2020avatar} underline a lower popularity of the appearance factor compared to the control and point of view, when assessing in combination (preference of choosing first the degree of control, before the degree of appearance). This latter conclusion might not be easily extrapolated to the case of a video-based self-avatar, as there is no need of avatar control when moving or interacting with real objects (only with virtual ones).

Improvement on presence and embodiment perceived in this work are in line with our previous work where virtual hands (using controllers) and real hands conditions were compared \cite{villegas2020realistic}, showing higher difference between the chroma-based and pure VR conditions. This is due to the fact, that in the present study all conditions share the same hand tracking algorithm, while in \cite{villegas2020realistic} the tracking in the VR condition was done through the VR controllers.

Our results are not in line with the conclusions extracted in \cite{gisbergen2020real}, where users needed to walk on a $60$m high broken pathway under two different conditions: $i)$ high resemblance shoes, or generic shoes. Authors stated that under extreme situations which trigger psychological arousal such as stress, high avatar-owner resemblance is not a requirement. 

\subsection{Benefits of Video-based Self-avatar}

The use of video-based self-avatar, provides some important benefits, when compared with traditional avatars. First, there is no need to scale the avatar to match user's body dimensions, as segmentation is done using a $1:1$ scale, allowing the user to have an accurate spatial perception. Also, there are no problems related to misalignment between virtual and real body \cite{ogawa2020effect}, uncanny valley effect \cite{bhargava2021did}, or self-avatar follower effect \cite{gonzalez2020self}. Extending egocentric segmentation to other items beyond human body parts, would allow the user to easily interact with real objects while immersed. Last but not least, users recognize the avatars as their own body with high acceptance values.

\subsection{Drawbacks of Video-based Self-avatars}

There are still certain factors that can be improved for a better integration of video-based self-avatars in MR application. One clear issue is the discrepancy of illumination conditions between the video-based self-avatar and the immersive environment where it is integrated (e.g. 360 video, VR environment). It would be also desirable to further investigate how to discard false positives from the segmentation algorithm, so that they do not downgrade the user experience, e.g. by using also depth. 
 
\subsection{Applications of  Video-based Self-avatars}

For evaluation purposes, the technology of video-based self-avatar has been presented in this work integrated in a gamified immersive experience. However, we believe it can play an important role in many emerging use cases, namely:
\begin{itemize}
    \item seeing your own body and others peers with high fidelity \cite{joachimczak2022creating} on immersive telepresence systems \cite{kachach2021owl,arora2022augmenting} or other type of immersive communication systems \cite{perez2022emerging} can further foster SoP, co-presence and communication skills.
    \item seeing your own body and being able to seamlessly interact with real object on immersive training purposes \cite{cortes2022qoe}.
    \item allowing, using the same segmentation technology, to include other relevant objects from the user's local reality such as a notebook or smartphone. 
\end{itemize}

\subsection{Limitations}

One of the limitation of this study is related to the lack of comparison with a full body avatar condition due to the lack of full body commercial tracking solutions. Also, the conclusions extracted from our user study are extracted from questionnaires answered by a population subset, which, although being balanced in terms of gender, lacks other diversity factors such as height, age, skin colour, among others. 

%


\section{Conclusions}
\label{sec:conclu}

In this work, we have thoroughly described the state of the art of user representation for MR applications, including a wide range of solutions such as realistic full-body avatars, virtual hands only, or video-based full-body representation. We have highlighted their advantages and disadvantages, focusing specially on the users perception of presence and embodiment. Besides, we reviewed specific state of the art implementations of video-based egocentric segmentation solutions for MR applications, highlighting the need of a more accurate and fast segmentation algorithm. To the best of our knowledge, we could not find any E2E implementation which allows the usage of deep learning based segmentation solutions on commercial wireless VR devices. 

Furthermore, we have presented our E2E system which provides full body video-based self-avatar user representation for MR applications. The E2E can be divided in three different modules: video pass-through solution for the Meta Quest 2, our real-time egocentric body segmentation algorithm, and our optimized offloading architecture. All the different components and implementation details have been described in detail. The E2E system has been evaluated using an arbitrary hardware setup. The achieved results showed the implementation's capability to fulfil the real-time requirements, providing update rates above 60 Hz. Finally we described our delay-correction buffers to overcome the round trip latencies above the update rate of 60 Hz ensuring an accurate alignment between the color and received mask frames. 

We have also presented the design of our subjective evaluation user study. To the best of our knowledge, this is the first user study which aims to evaluate the effect of video-based self-avatar user representation. We used a custom gamified experience, thoroughly described in this paper. The goal was to evaluate and compare three different user representation solutions: virtual hands, chroma-based full-body segmentation, and deep-learning based full-body segmentation. The questionnaire, to be filled by 58 participants, aimed to evaluate the embodiment, presence and visual quality. The results showed moderate improvements of the video-based segmentation algorithms (chroma and deep learning) with respect to the virtual hands solution, specially in terms of location, appearance and response. Within-subject embodiment analysis showed slight improvements of the deep-learning solution with respect to the chroma. Presence results where high with negligible differences between the algorithms. In terms of visual quality, the greatest differences are in the perception of the user's legs and the false-positives, as expected. We observed an increment of around one quality level of the deep-learning algorithm with respect to the chroma solution. We also measured the overall acceptability of the three solutions, which showed extremely high values with and no significant differences. Finally, we thoroughly analyzed and discussed the insights obtained from the free text comments given by the users.

The overall results allowed us to reach two main conclusions: $i)$ video-avatars provided marginal improvement only in certain components of the perceived embodiment compared to the virtual hands solution, $ii)$ deep-learning solution offers better visual quality and perception of ownership and appearance than the chroma solution. In general we observed low differences between the three conditions. due to the high levels of sense of embodiment and presence obtained. The observed low differences between the conditions might by due to the fact that in all three conditions the same robust and accurate hand tracking algorithm provided by the Meta Quest 2. Besides, the proposed E2E video-based solution is at the edge of the state of the art and consequently, it lacks the maturity provided by the commercial virtual-hands solution. However, the presented E2E system is already useful in its current state, and can be consider the starting point for future MR distributed implementations and potential improvements. We also observed that the scores obtained for presence or embodiment are high for all conditions, making it difficult to discriminate among them. Additional Quality of Experience factors should be explored to study differences between MR technologies.

Finally, we discussed some advantages and disadvantages of video-based self-avatars solutions and potential applications. We believe this solution can relevant in applications like telepresence, training and education. 

In the near future, we would like to explore other evaluation methods to extend the obtained results and reach clearer insights. Besides, we believe it would be interesting to repeat the experiments with a more balanced set of users in terms of age, skin colour, social background, among others. One of the limitations of our results is the fact that the video-based solutions are not compared to full-body virtual realistic avatars as full-body tracking is still not commercially available. It will be interesting to perform this comparison in the future. 


%

\appendices


\ifCLASSOPTIONcompsoc
  \section*{Acknowledgments}
\else
  \section*{Acknowledgment}
\fi

The authors would like to thank all the people conducting the user's study.

This work has received funding from the European Union (EU) Horizon 2020 research and innovation programme under the Marie Skłodowska-Curie ETN TeamUp5G, grant agreement No. 813391.

\ifCLASSOPTIONcaptionsoff
  \newpage
\fi



\bibliographystyle{IEEEtran}
\bibliography{bare_jrnl_compsoc}

\begin{thebibliography}{10}
\providecommand{\url}[1]{#1}
\csname url@samestyle\endcsname
\providecommand{\newblock}{\relax}
\providecommand{\bibinfo}[2]{#2}
\providecommand{\BIBentrySTDinterwordspacing}{\spaceskip=0pt\relax}
\providecommand{\BIBentryALTinterwordstretchfactor}{4}
\providecommand{\BIBentryALTinterwordspacing}{\spaceskip=\fontdimen2\font plus
\BIBentryALTinterwordstretchfactor\fontdimen3\font minus
  \fontdimen4\font\relax}
\providecommand{\BIBforeignlanguage}[2]{{%
\expandafter\ifx\csname l@#1\endcsname\relax
\typeout{** WARNING: IEEEtran.bst: No hyphenation pattern has been}%
\typeout{** loaded for the language `#1'. Using the pattern for}%
\typeout{** the default language instead.}%
\else
\language=\csname l@#1\endcsname
\fi
#2}}
\providecommand{\BIBdecl}{\relax}
\BIBdecl

\bibitem{lok2003effects}
B.~Lok, S.~Naik, M.~Whitton, and F.~P. Brooks, ``Effects of handling real
  objects and self-avatar fidelity on cognitive task performance and sense of
  presence in virtual environments,'' \emph{Presence}, vol.~12, no.~6, pp.
  615--628, 2003.

\bibitem{slater1993influence}
M.~Slater and M.~Usoh, ``The influence of a virtual body on presence in
  immersive virtual environments,'' in \emph{Proc. of VR}, 1993, pp. 34--42.

\bibitem{mcmanus2011influence}
E.~McManus, B.~Bodenheimer, S.~Streuber, S.~De~La~Rosa, H.~H. B{\"u}lthoff, and
  B.~J. Mohler, ``The influence of avatar (self and character) animations on
  distance estimation, object interaction and locomotion in immersive virtual
  environments,'' in \emph{Proc. of the ACM SIGGRAPH SAP}, 2011, pp. 37--44.

\bibitem{ebrahimi2018investigating}
E.~Ebrahimi, L.~S. Hartman, A.~Robb, C.~C. Pagano, and S.~V. Babu,
  ``Investigating the effects of anthropomorphic fidelity of self-avatars on
  near field depth perception in immersive virtual environments,'' in
  \emph{2018 IEEE conference on virtual reality and 3D user interfaces
  (VR)}.\hskip 1em plus 0.5em minus 0.4em\relax IEEE, 2018, pp. 1--8.

\bibitem{steed2016impactselfavatar}
A.~Steed, Y.~Pan, F.~Zisch, and W.~Steptoe, ``The impact of a self-avatar on
  cognitive load in immersive virtual reality,'' in \emph{2016 IEEE Virtual
  Reality (VR)}, 2016, pp. 67--76.

\bibitem{pan2017impact}
Y.~Pan and A.~Steed, ``The impact of self-avatars on trust and collaboration in
  shared virtual environments,'' \emph{PloS one}, vol.~12, no.~12, p. e0189078,
  2017.

\bibitem{argelaguet2016role}
F.~Argelaguet, L.~Hoyet, M.~Trico, and A.~L{\'e}cuyer, ``The role of
  interaction in virtual embodiment: Effects of the virtual hand
  representation,'' in \emph{Proc. IEEE VR}, 2016, pp. 3--10.

\bibitem{fribourg2020avatar}
R.~Fribourg, F.~Argelaguet, A.~L{\'e}cuyer, and L.~Hoyet, ``Avatar and sense of
  embodiment: Studying the relative preference between appearance, control and
  point of view,'' \emph{IEEE transactions on visualization and computer
  graphics}, vol.~26, no.~5, pp. 2062--2072, 2020.

\bibitem{kilteni2012sense}
K.~Kilteni, R.~Groten, and M.~Slater, ``The sense of embodiment in virtual
  reality,'' \emph{Presence: Teleoperators and Virtual Environments}, vol.~21,
  no.~4, pp. 373--387, 2012.

\bibitem{dodds2011talk}
T.~J. Dodds, B.~J. Mohler, and H.~H. B{\"u}lthoff, ``Talk to the virtual hands:
  Self-animated avatars improve communication in head-mounted display virtual
  environments,'' \emph{PloS one}, vol.~6, no.~10, p. e25759, 2011.

\bibitem{gruosso2021}
M.~Gruosso, N.~Capece, and U.~Erra, ``Exploring upper limb segmentation with
  deep learning for augmented virtuality,'' 2021.

\bibitem{ogawa2020you}
N.~Ogawa, T.~Narumi, H.~Kuzuoka, and M.~Hirose, ``Do you feel like passing
  through walls?: Effect of self-avatar appearance on facilitating realistic
  behavior in virtual environments,'' in \emph{Proceedings of the 2020 CHI
  Conference on Human Factors in Computing Systems}, 2020, pp. 1--14.

\bibitem{dewez2019ismar}
D.~Dewez, R.~Fribourg, F.~Argelaguet, L.~Hoyet, D.~Mestre, M.~Slater, and
  A.~Lécuyer, ``Influence of personality traits and body awareness on the
  sense of embodiment in virtual reality,'' in \emph{2019 IEEE International
  Symposium on Mixed and Augmented Reality (ISMAR)}, 2019, pp. 123--134.

\bibitem{gonzalez2020rocketbox}
M.~Gonzalez-Franco, E.~Ofek, Y.~Pan, A.~Antley, A.~Steed, B.~Spanlang,
  A.~Maselli, D.~Banakou, N.~Pelechano, S.~Orts-Escolano \emph{et~al.}, ``The
  rocketbox library and the utility of freely available rigged avatars,''
  \emph{Frontiers in virtual reality}, p.~20, 2020.

\bibitem{loper2015smpl}
M.~Loper, N.~Mahmood, J.~Romero, G.~Pons-Moll, and M.~J. Black, ``Smpl: A
  skinned multi-person linear model,'' \emph{ACM transactions on graphics
  (TOG)}, vol.~34, no.~6, pp. 1--16, 2015.

\bibitem{thaler2018visual}
A.~Thaler, I.~Piryankova, J.~K. Stefanucci, S.~Pujades, S.~de~La~Rosa,
  S.~Streuber, J.~Romero, M.~J. Black, and B.~J. Mohler, ``Visual perception
  and evaluation of photo-realistic self-avatars from 3d body scans in males
  and females,'' \emph{Frontiers in ICT}, p.~18, 2018.

\bibitem{ogawa2019virtual}
N.~Ogawa, T.~Narumi, and M.~Hirose, ``Virtual hand realism affects object size
  perception in body-based scaling,'' in \emph{2019 IEEE Conference on Virtual
  Reality and 3D User Interfaces (VR)}.\hskip 1em plus 0.5em minus 0.4em\relax
  IEEE, 2019, pp. 519--528.

\bibitem{gorisse2019robot}
G.~Gorisse, O.~Christmann, S.~Houzangbe, and S.~Richir, ``From robot to virtual
  doppelganger: Impact of visual fidelity of avatars controlled in third-person
  perspective on embodiment and behavior in immersive virtual environments,''
  \emph{Frontiers in Robotics and AI}, vol.~6, p.~8, 2019.

\bibitem{waltemate2018impact}
T.~Waltemate, D.~Gall, D.~Roth, M.~Botsch, and M.~E. Latoschik, ``The impact of
  avatar personalization and immersion on virtual body ownership, presence, and
  emotional response,'' \emph{IEEE transactions on visualization and computer
  graphics}, vol.~24, no.~4, pp. 1643--1652, 2018.

\bibitem{yu2021avatars}
K.~Yu, G.~Gorbachev, U.~Eck, F.~Pankratz, N.~Navab, and D.~Roth, ``Avatars for
  teleconsultation: effects of avatar embodiment techniques on user perception
  in 3d asymmetric telepresence,'' \emph{IEEE Transactions on Visualization and
  Computer Graphics}, vol.~27, no.~11, pp. 4129--4139, 2021.

\bibitem{pan2019foot}
Y.~Pan and A.~Steed, ``How foot tracking matters: The impact of an animated
  self-avatar on interaction, embodiment and presence in shared virtual
  environments,'' \emph{Frontiers in Robotics and AI}, p. 104, 2019.

\bibitem{serra2020natural}
S.~Serra, R.~Kachach, E.~Gonzalez-Sosa, and A.~Villegas, ``Natural user
  interfaces for mixed reality: Controlling virtual objects with your real
  hands,'' in \emph{2020 IEEE Conference on Virtual Reality and 3D User
  Interfaces Abstracts and Workshops (VRW)}.\hskip 1em plus 0.5em minus
  0.4em\relax IEEE, 2020, pp. 712--713.

\bibitem{bonfert2022kicking}
M.~Bonfert, S.~Lemke, R.~Porzel, and R.~Malaka, ``Kicking in virtual reality:
  The influence of foot visibility on the shooting experience and accuracy,''
  in \emph{2022 IEEE Conference on Virtual Reality and 3D User Interfaces
  (VR)}.\hskip 1em plus 0.5em minus 0.4em\relax IEEE, 2022, pp. 711--718.

\bibitem{bozgeyikli2022tangiball}
L.~L. Bozgeyikli and E.~Bozgeyikli, ``Tangiball: Foot-enabled embodied tangible
  interaction with a ball in virtual reality,'' in \emph{2022 IEEE Conference
  on Virtual Reality and 3D User Interfaces (VR)}.\hskip 1em plus 0.5em minus
  0.4em\relax IEEE, 2022, pp. 812--820.

\bibitem{xu2019towards}
C.~Xu, J.~He, X.~Zhang, X.~Zhou, and S.~Duan, ``Towards human motion tracking:
  Multi-sensory imu/toa fusion method and fundamental limits,''
  \emph{Electronics}, vol.~8, no.~2, p. 142, 2019.

\bibitem{jayaraj2017improving}
L.~Jayaraj, J.~Wood, and M.~Gibson, ``Improving the immersion in virtual
  reality with real-time avatar and haptic feedback in a cricket simulation,''
  in \emph{2017 IEEE international symposium on mixed and augmented reality
  (ISMAR-adjunct)}.\hskip 1em plus 0.5em minus 0.4em\relax IEEE, 2017, pp.
  310--314.

\bibitem{gonzalez2020movebox}
M.~Gonzalez-Franco, Z.~Egan, M.~Peachey, A.~Antley, T.~Randhavane, P.~Panda,
  Y.~Zhang, C.~Y. Wang, D.~F. Reilly, T.~C. Peck \emph{et~al.}, ``Movebox:
  Democratizing mocap for the microsoft rocketbox avatar library,'' in
  \emph{2020 IEEE International Conference on Artificial Intelligence and
  Virtual Reality (AIVR)}.\hskip 1em plus 0.5em minus 0.4em\relax IEEE, 2020,
  pp. 91--98.

\bibitem{fiore2012towards}
L.~P. Fiore and V.~Interrante, ``Towards achieving robust video selfavatars
  under flexible environment conditions,'' \emph{International Journal of VR},
  vol.~11, no.~3, pp. 33--41, 2012.

\bibitem{bruder2009enhancing}
G.~Bruder, F.~Steinicke, K.~Rothaus, and K.~Hinrichs, ``Enhancing presence in
  head-mounted display environments by visual body feedback using head-mounted
  cameras,'' in \emph{Proc. Int. Conf. on CW}, 2009, pp. 43--50.

\bibitem{immersirve_gastronomic2019}
P.~Perez, E.~Gonzalez-Sosa, R.~Kachach, J.~Ruiz, F.~Pereira, and A.~Villegas,
  ``Immersive gastronomic experience with distributed reality,'' in \emph{Proc
  . of IEEE WEVR}, 2019, pp. 1--4.

\bibitem{gunther2015aughanded}
T.~G{\"u}nther, I.~S. Franke, and R.~Groh, ``Aughanded virtuality-the hands in
  the virtual environment,'' in \emph{Proc. of IEEE 3DUI}, 2015, pp. 157--158.

\bibitem{rauter2019augmenting}
M.~Rauter, C.~Abseher, and M.~Safar, ``Augmenting virtual reality with near
  real world objects,'' in \emph{Proc. IEEE VR}, 2019, pp. 1134--1135.

\bibitem{lee2016enhancing}
G.~A. Lee, J.~Chen, M.~Billinghurst, and R.~Lindeman, ``Enhancing immersive
  cinematic experience with augmented virtuality,'' in \emph{IEEE International
  Symposium on Mixed and Augmented Reality}, 2016, pp. 115--116.

\bibitem{alaee2018user}
G.~Alaee, A.~P. Deasi, L.~Pena-Castillo, E.~Brown, and O.~Meruvia-Pastor, ``A
  user study on augmented virtuality using depth sensing cameras for near-range
  awareness in immersive vr,'' in \emph{IEEE VR’s 4th Workshop on Everyday
  Virtual Reality (WEVR 2018)}, vol.~10, 2018, p.~3.

\bibitem{gonzalez2020enhanced}
E.~Gonzalez-Sosa, P.~Perez, R.~Tolosana, R.~Kachach, and A.~Villegas,
  ``Enhanced self-perception in mixed reality: Egocentric arm segmentation and
  database with automatic labeling,'' \emph{IEEE Access}, vol.~8, pp.
  146\,887--146\,900, 2020.

\bibitem{gonzalez2022_realtimeseg}
\BIBentryALTinterwordspacing
E.~Gonzalez-Sosa, A.~Gajic, D.~Gonzalez-Morin, G.~Robledo, P.~Perez, and
  A.~Villegas, ``Real time egocentric segmentation for video-self avatar in
  mixed reality,'' 2022. [Online]. Available:
  \url{https://arxiv.org/abs/2207.01296}
\BIBentrySTDinterwordspacing

\bibitem{xiang2019thundernet}
W.~Xiang, H.~Mao, and V.~Athitsos, ``Thundernet: A turbo unified network for
  real-time semantic segmentation,'' in \emph{2019 IEEE winter conference on
  applications of computer vision (WACV)}.\hskip 1em plus 0.5em minus
  0.4em\relax IEEE, 2019, pp. 1789--1796.

\bibitem{gonzalez2022bringing}
D.~Gonzalez-Morin, E.~Gonzalez-Sosa, P.~Perez-Garcia, and A.~Villegas,
  ``Bringing real body as self-avatar into mixed reality: A gamified volcano
  experience,'' in \emph{2022 IEEE Conference on Virtual Reality and 3D User
  Interfaces Abstracts and Workshops (VRW)}, 2022, pp. 794--795.

\bibitem{morin2022democratic}
D.~G. Mor{\'\i}n, F.~Pereira, E.~Gonz{\'a}lez, P.~P{\'e}rez, and A.~Villegas,
  ``Democratic video pass-through for commercial virtual reality devices,'' in
  \emph{2022 IEEE Conference on Virtual Reality and 3D User Interfaces
  Abstracts and Workshops (VRW)}.\hskip 1em plus 0.5em minus 0.4em\relax IEEE,
  2022, pp. 790--791.

\bibitem{chen2017effect}
G.~Chen, J .and~Lee, M.~Billinghurst, R.~W. Lindeman, and C.~Bartneck, ``The
  effect of user embodiment in av cinematic experience,'' in \emph{Proc of
  ICAT-EGVE}, 2017.

\bibitem{pigny2019using}
P.-O. Pigny and L.~Dominjon, ``Using cnns for users segmentation in video
  see-through augmented virtuality,'' in \emph{2019 IEEE International
  Conference on Artificial Intelligence and Virtual Reality (AIVR)}, 2019, pp.
  229--2295.

\bibitem{bai2021bringing}
H.~Bai, L.~Zhang, J.~Yang, and M.~Billinghurst, ``Bringing full-featured mobile
  phone interaction into virtual reality,'' \emph{Computers \& Graphics},
  vol.~97, pp. 42--53, 2021.

\bibitem{cortes2022qoe}
C.~Cort{\'e}s, M.~Rubio, P.~P{\'e}rez, B.~S{\'a}nchez, and N.~Garc{\'\i}a,
  ``Qoe study of natural interaction in extended reality environment for
  immersive training,'' in \emph{2022 IEEE Conference on Virtual Reality and 3D
  User Interfaces Abstracts and Workshops (VRW)}.\hskip 1em plus 0.5em minus
  0.4em\relax IEEE, 2022, pp. 363--368.

\bibitem{tian2019enhancing}
Y.~Tian, C.-W. Fu, S.~Zhao, R.~Li, X.~Tang, X.~Hu, and P.-A. Heng, ``Enhancing
  augmented vr interaction via egocentric scene analysis,'' \emph{Proceedings
  of the ACM on Interactive, Mobile, Wearable and Ubiquitous Technologies},
  vol.~3, no.~3, pp. 1--24, 2019.

\bibitem{roxas2018occlusion}
M.~Roxas, T.~Hori, T.~Fukiage, Y.~Okamoto, and T.~Oishi, ``Occlusion handling
  using semantic segmentation and visibility-based rendering for mixed
  reality,'' in \emph{Proceedings of the 24th ACM Symposium on Virtual Reality
  Software and Technology}, 2018, pp. 1--8.

\bibitem{libuvc}
\BIBentryALTinterwordspacing
{libuvc: a Cross-Platform Library for USB Video Devices}. (2021, Dec).
  [Online]. Available: \url{https://ken.tossell.net/libuvc/doc/}
\BIBentrySTDinterwordspacing

\bibitem{CameraCalibration}
Z.~Zhang, ``A flexible new technique for camera calibration,'' \emph{IEEE
  Transactions on Pattern Analysis and Machine Intelligence}, vol.~22, no.~11,
  pp. 1330--1334, 2000.

\bibitem{DistortionEquations}
D.~C. Brown, ``Close-range camera calibration,'' \emph{Photogramm. Eng.},
  vol.~37, 12 2002.

\bibitem{CameraIMU}
F.~M. Mirzaei and S.~I. Roumeliotis, ``A kalman filter-based algorithm for
  imu-camera calibration: Observability analysis and performance evaluation,''
  \emph{IEEE Transactions on Robotics}, vol.~24, no.~5, pp. 1143--1156, 2008.

\bibitem{Aruco}
S.~Garrido-Jurado \emph{et~al.}, ``Automatic generation and detection of highly
  reliable fiducial markers under occlusion,'' \emph{Pattern Recognition},
  vol.~47, no.~6, pp. 2280 -- 2292, 2014.

\bibitem{guo2018review}
Y.~Guo, Y.~Liu, T.~Georgiou, and M.~S. Lew, ``A review of semantic segmentation
  using deep neural networks,'' \emph{International journal of multimedia
  information retrieval}, vol.~7, no.~2, pp. 87--93, 2018.

\bibitem{ZMQBench}
P.~Sommer, F.~Schellroth, M.~Fischer, and J.~Schlechtendahl, ``Message-oriented
  middleware for industrial production systems,'' in \emph{2018 IEEE 14th
  International Conference on Automation Science and Engineering (CASE)}.\hskip
  1em plus 0.5em minus 0.4em\relax Munich, Germany: IEEE, 2018, pp. 1217--1223.

\bibitem{imxoffloading}
\BIBentryALTinterwordspacing
D.~Gonz\'{a}lez~Mor\'{\i}n, M.~J. L\'{o}pez~Morales, P.~P\'{e}rez, and
  A.~Villegas, ``Tcp-based distributed offloading architecture for the future
  of untethered immersive experiences in wireless networks,'' in \emph{ACM
  International Conference on Interactive Media Experiences}, ser. IMX
  '22.\hskip 1em plus 0.5em minus 0.4em\relax New York, NY, USA: Association
  for Computing Machinery, 2022, p. 121–132. [Online]. Available:
  \url{https://doi.org/10.1145/3505284.3529963}
\BIBentrySTDinterwordspacing

\bibitem{gonzalez2018avatar}
M.~Gonzalez-Franco and T.~C. Peck, ``Avatar embodiment. towards a standardized
  questionnaire,'' \emph{Frontiers in Robotics and AI}, vol.~5, p.~74, 2018.

\bibitem{witmer2005factor}
B.~G. Witmer, C.~J. Jerome, and M.~J. Singer, ``The factor structure of the
  presence questionnaire,'' \emph{Presence: Teleoperators \& Virtual
  Environments}, vol.~14, no.~3, pp. 298--312, 2005.

\bibitem{perez2021ecological}
P.~P{\'e}rez, E.~Gonz{\'a}lez-Sosa, R.~Kachach, F.~Pereira, and
  {\'A}.~Villegas, ``Ecological validity through gamification: an experiment
  with a mixed reality escape room,'' in \emph{2021 IEEE International
  Conference on Artificial Intelligence and Virtual Reality (AIVR)}.\hskip 1em
  plus 0.5em minus 0.4em\relax IEEE, 2021, pp. 179--183.

\bibitem{perez2018towards}
P.~P{\'e}rez, N.~Oyaga, J.~J. Ruiz, and {\'A}.~Villegas, ``Towards systematic
  analysis of cybersickness in high motion omnidirectional video,'' in
  \emph{2018 Tenth International Conference on Quality of Multimedia Experience
  (QoMEX)}.\hskip 1em plus 0.5em minus 0.4em\relax IEEE, 2018, pp. 1--3.

\bibitem{hossfeld2016qoe}
T.~Ho{\ss}feld, P.~E. Heegaard, M.~Varela, and S.~M{\"o}ller, ``Qoe beyond the
  mos: an in-depth look at qoe via better metrics and their relation to mos,''
  \emph{Quality and User Experience}, vol.~1, no.~1, pp. 1--23, 2016.

\bibitem{reichheld2003one}
F.~F. Reichheld, ``The one number you need to grow,'' \emph{Harvard business
  review}, vol.~81, no.~12, pp. 46--55, 2003.

\bibitem{villegas2020realistic}
A.~Villegas, P.~Perez, R.~Kachach, F.~Pereira, and E.~Gonzalez-Sosa,
  ``Realistic training in vr using physical manipulation,'' in \emph{2020 IEEE
  Conference on Virtual Reality and 3D User Interfaces Abstracts and Workshops
  (VRW)}, 2020, pp. 109--118.

\bibitem{narwaria2018data}
M.~Narwaria, L.~Krasula, and P.~Le~Callet, ``Data analysis in multimedia
  quality assessment: Revisiting the statistical tests,'' \emph{IEEE
  Transactions on Multimedia}, vol.~20, no.~8, pp. 2063--2072, 2018.

\bibitem{gisbergen2020real}
M.~S.~v. Gisbergen, I.~Sensagir, and J.~Relouw, ``How real do you see yourself
  in vr? the effect of user-avatar resemblance on virtual reality experiences
  and behaviour,'' in \emph{Augmented Reality and Virtual Reality}.\hskip 1em
  plus 0.5em minus 0.4em\relax Springer, 2020, pp. 401--409.

\bibitem{ogawa2020effect}
N.~Ogawa, T.~Narumi, and M.~Hirose, ``Effect of avatar appearance on detection
  thresholds for remapped hand movements,'' \emph{IEEE TRANS. on VCG}, 2020.

\bibitem{bhargava2021did}
A.~Bhargava, R.~Venkatakrishnan, R.~Venkatakrishnan, H.~Solini, K.~M. Lucaites,
  A.~Robb, C.~Pagano, and S.~Babu, ``Did i hit the door effects of self-avatars
  and calibration in a person-plus-virtual-object system on perceived frontal
  passability in vr,'' \emph{IEEE Transactions on Visualization and Computer
  Graphics}, 2021.

\bibitem{gonzalez2020self}
M.~Gonzalez-Franco, B.~Cohn, E.~Ofek, D.~Burin, and A.~Maselli, ``The
  self-avatar follower effect in virtual reality,'' in \emph{2020 IEEE
  Conference on Virtual Reality and 3D User Interfaces (VR)}.\hskip 1em plus
  0.5em minus 0.4em\relax IEEE, 2020, pp. 18--25.

\bibitem{joachimczak2022creating}
M.~Joachimczak, J.~Liu, and H.~Ando, ``Creating 3d personal avatars with high
  quality facial expressions for telecommunication and telepresence,'' in
  \emph{2022 IEEE Conference on Virtual Reality and 3D User Interfaces
  Abstracts and Workshops (VRW)}.\hskip 1em plus 0.5em minus 0.4em\relax IEEE,
  2022, pp. 856--857.

\bibitem{kachach2021owl}
R.~Kachach, S.~Morcuende, D.~Gonzalez-Morin, P.~Perez-Garcia, E.~Gonzalez-Sosa,
  F.~Pereira, and A.~Villegas, ``The owl: Immersive telepresence communication
  for hybrid conferences,'' in \emph{2021 IEEE International Symposium on Mixed
  and Augmented Reality Adjunct (ISMAR-Adjunct)}.\hskip 1em plus 0.5em minus
  0.4em\relax IEEE, 2021, pp. 451--452.

\bibitem{arora2022augmenting}
N.~Arora, M.~Suomalainen, M.~Pouke, E.~G. Center, K.~J. Mimnaugh, A.~P.
  Chambers, S.~Pouke, and S.~M. LaValle, ``Augmenting immersive telepresence
  experience with a virtual body,'' \emph{arXiv preprint arXiv:2202.00900},
  2022.

\bibitem{perez2022emerging}
P.~Pérez, E.~Gonzalez-Sosa, J.~Gutiérrez, and N.~García.

\end{thebibliography}
%

%


\begin{IEEEbiography}[{\includegraphics[width=1in,height=1.25in,clip,keepaspectratio]{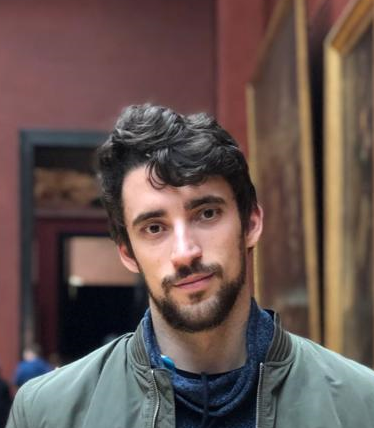}}]
{Diego Gonzalez} is a Ph.D student at Nokia Extended Reality Lab, enrolled with Universidad Carlos III de Madrid, Spain. He received his B.Sc. and M.Sc. in industrial engineering from Universidad Politécnica de Madrid in 2015 and 2018 respectively. In 2018, he received his M.Sc. in systems, control and robotics from Kunliga Tekniska Hgskolan (KTH), Stockholm, Sweden. After receiving his M.Sc. degrees, he joined Ericsson Research's Devices Technologies group as a researcher, where his research interest focused on augmented reality technologies, a field in which he holds three patents. In August 2019, he joined Nokia Bell Labs as a Ph.D. student. He is currently pursuing a Ph.D. focused on the application of ultra-dense networks for the implementation of distributed media rendering.
\end{IEEEbiography}

\begin{IEEEbiography}[{\includegraphics[width=1in,height=1.25in,clip,keepaspectratio]{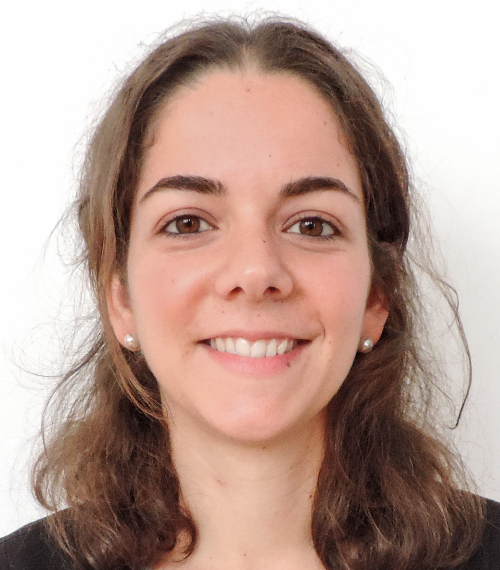}}]
{Ester Gonzalez-Sosa}received the B.S. in computer science and M.Sc in electrical engineering from Universidad de Las Palmas de Gran Canaria in $2012$ and $2014$, respectively. In June $2017$ she obtained her PhD degree from Universidad Autonoma de Madrid, within the Biometrics and Data Pattern Analytics - BiDA Lab. In October 2017 she joined Nokia Bell Labs as a researcher in future video technologies. At the moment, she works in the Extended Reality Lab in Nokia, where she focuses on computer vision applied to Mixed Reality applications, with a focus on real-time, performance in the wild, and application related to fostering human communications. She has carried out several research internships in worldwide leading groups in biometric recognition such as TNO, EURECOM, or Rutgers University. Gonzalez-Sosa has been the recipient of the competitive Obra Social La CAIXA Scholarship ($2012$), the UNITECO AWARD from the Spanish Association of Electrical Engineers ($2013$) and the European Biometrics Research Award ($2018$).
\end{IEEEbiography}

\begin{IEEEbiography}[{\includegraphics[width=1in,height=1.25in,clip,keepaspectratio]{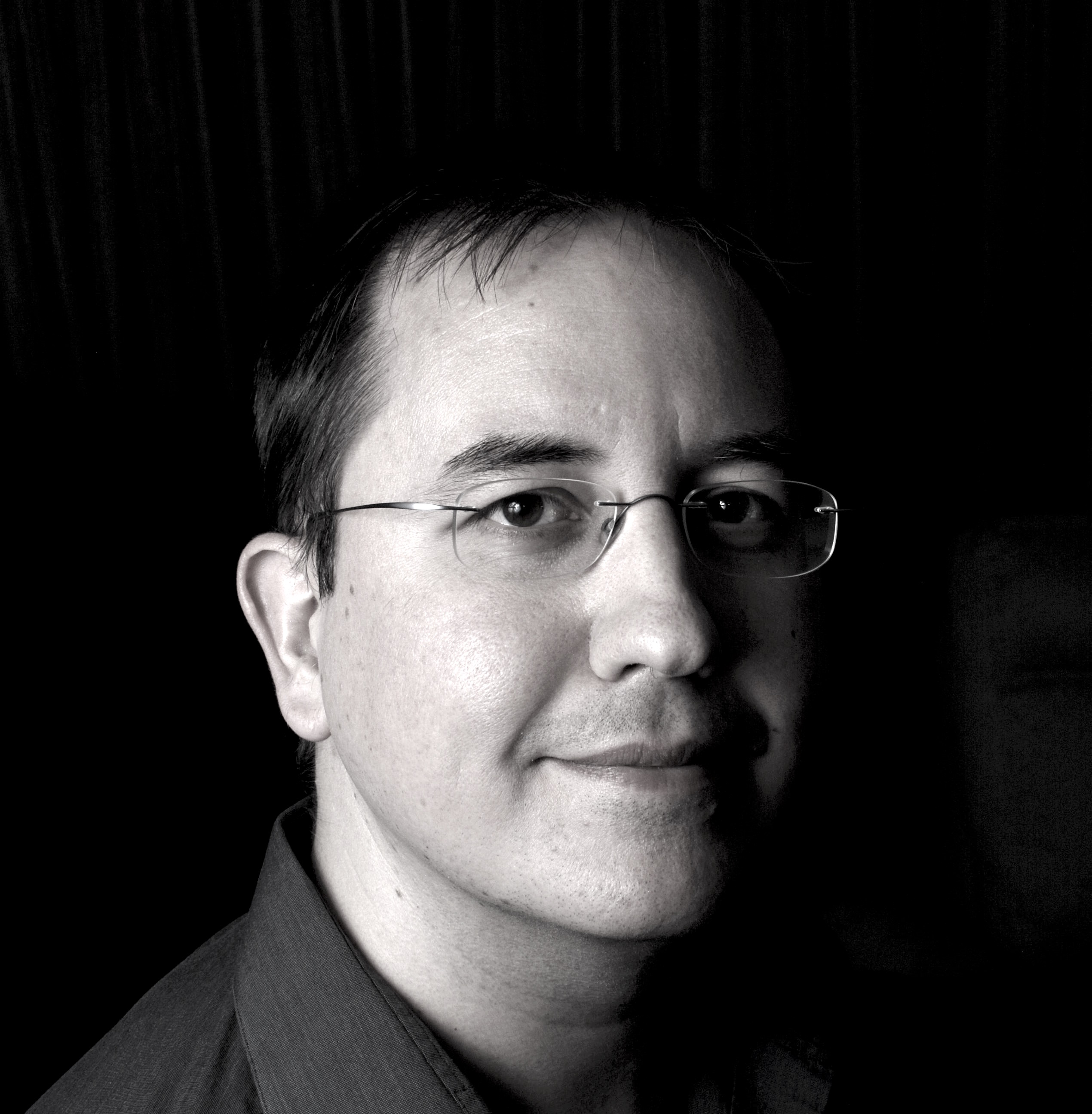}}]
{Pablo Perez} is Lead Scientist at Nokia Extended Reality Lab (Madrid, Spain). He is Telecommunication Enginner (BSc and MSs, 2004) and PhD in Telecommunications (2013) from Universidad Politecnica de Madrid, Spain, and Nokia Distinguished Member of Technical Staff (2022) . He has worked as $R\&D$ engineer of digital video products and services in Telefonica, Alcatel-Lucent and Nokia; as well as a researcher in future video technologies in Nokia Bell Labs. He is currently leading the scientific activities of Nokia XR Lab, addressing the end-to-end technological chain of the use of Extended Reality for human communication: networking, system architecture, processing algorithms, quality of experience and human-computer interaction.
\end{IEEEbiography}

\begin{IEEEbiography}[{\includegraphics[width=1in,height=1.25in,clip,keepaspectratio]{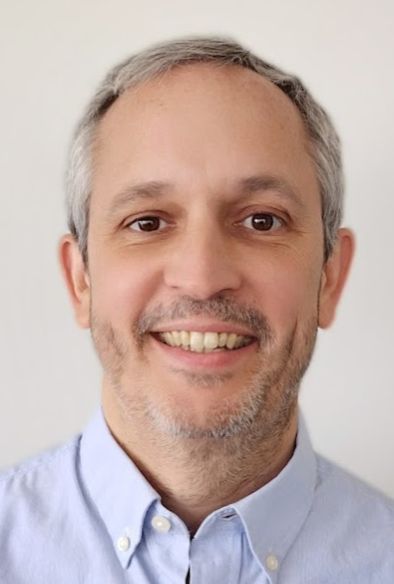}}]%
{Alvaro Villegas} leads the Extended Reality Lab in Nokia, a research center focused in the application of immersive media (VR, AR, XR) to human communications. He received a six-year telecommunications engineering degree at Universidad Politécnica de Madrid (Spain) and he completed an MBA Core Program at ESCP Europe Business School. Alvaro received the Distinguished Member of Technical Staff title from Bell Labs. He has dedicated his nearly 30 years of professional life to innovate in digital video in different companies: Telefónica, ONO, Motorola, Nagravision, Alcatel-Lucent and Nokia, where he has filed more than 40 patents. In his former role as Head of Bell Labs in Nokia Spain and now as lead of XR Lab he applies XR, AI/ML and 5G/6G technologies to improve human communications.
\end{IEEEbiography}




\end{document}